# OmniCyRisk: Reliability-oriented Multi-Consequence Forecasting of Tropical Cyclone Impacts from Limited and Partially Observed Evidence

**Yongchuan Yang[1], Naiyu Wang[1,*], Yongliang Shen[2], Weiming Lu[3]**

[1] College of Civil Engineering and Architecture, Zhejiang University, Hangzhou, China
[2] College of Software Engineering, Zhejiang University, Hangzhou, China
[3] College of Artificial Intelligence, Zhejiang University, Hangzhou, China

*Corresponding Author: naiyuwang@zju.edu.cn

**ABSTRACT:** Tropical cyclone impact records are scarce at the event level and fragmented across consequences and locations, yet each new report may contain information about impacts that remain unobserved. We developed OmniCyRisk to test whether such partial evidence can support coordinated forecasts of affected population, affected cropland, damaged buildings, and direct economic loss. The framework combines shared spatial features and consequence-adaptive heads with exposure-consistent county supervision, simulation-assisted initialization, and masked updating. In event-blocked tests of 20 cyclones and 90 counties in Zhejiang Province, China, OmniCyRisk achieved reliability comparable to architecture-matched single-task models (macro reliability 0.685 versus 0.675), although damaged-building performance declined. Its predictions more closely reproduced five of six observed cross-consequence rank relationships, with 59.4% lower mean pairwise correlation error as a descriptive measure. Simulation-assisted initialization improved 19 of 20 consequence–fold comparisons. Retrospective updates improved all four withheld consequences and yielded positive skill point estimates for all 11 non-reporting cities within Zhejiang; across years, pooled skill was positive but two of seven cyclones showed negative transfer. Scarcity is therefore also an evidence-organization problem: related reports can inform still-unreported impacts and places, provided updates are checked for negative transfer and retained performance. This creates a testable route toward multi-sector situational awareness, pending independent-region and prospective validation.

## 1. Introduction

Independent observations of damaging tropical cyclones accumulate slowly. The 1,800 county–event records for each consequence in this study arise from only 20 storms, and impacts are reported at administrative rather than grid scale. Records also differ across consequences, places, and reporting systems. Forecasting under these conditions depends on how existing evidence is combined, not simply on the nominal number of rows in an archive.

A cyclone affects people, agriculture, buildings, infrastructure, and economic activity through overlapping hazard, exposure, and vulnerability conditions. Each recorded consequence is therefore a partial observation of a common disaster process.

Building damage carries information about realized hazard intensity, built exposure, and regional vulnerability that may also be relevant to population impact or economic loss; agricultural impacts reveal complementary rainfall, land-use, and exposure patterns. These consequences are related but not interchangeable because their exposure bases, statistical distributions, and response mechanisms differ. Modeling them independently preserves specialization, but prevents evidence in one part of the impact system from informing another. We therefore formulate multi-consequence forecasting as an evidence-utilization problem: how can fragmented observations be coordinated to enhance the reliability and enable conditional transfer of forecasts across the wider impact system without erasing consequence-specific behavior?

This question is operationally consequential. Impact-based forecasting moves beyond hazard magnitude, location, and timing to estimate decision-relevant consequences (World Meteorological Organization, 2015), consistent with a broader risk perspective in which disaster impacts emerge from interacting hazard, exposure, and vulnerability conditions (Merz et al., 2021). A tropical cyclone does not produce impacts as a collection of independent outcomes, and decisions concerning evacuation, resource allocation, infrastructure protection, agricultural assistance, and recovery rarely depend on one indicator alone. They require a coherent picture of simultaneous human, agricultural, built-environment, and economic impacts. From the perspectives of reliability engineering and system safety, tropical cyclone impact forecasting is therefore a system-level risk-estimation problem under scarce, heterogeneous, and incompletely reported evidence. Such system-level impact information is relevant to emergency management agencies responsible for warning and response coordination, insurance and financial sectors assessing loss exposure, infrastructure operators planning protection strategies, and health and humanitarian organizations allocating recovery resources.

Most data-driven tropical cyclone impact studies predict one consequence at a time. Regional models address fine-resolution impacts or affected population (Lin and Wang, 2024; Meng et al., 2024); broader studies examine housing damage, displacement, historical losses, or affected-population fractions (Kooshki Forooshani et al., 2024; Kam et al., 2024; Vessey et al., 2026; Moss et al., 2026). Related work on road functionality, power outages, and flood damage also treats distinct responses (Wang and Wang, 2024; Yang et al., 2026; Xie et al., 2026). This specialization respects different impact mechanisms, but divides scarce event evidence among separate models. An outcome observed for one cyclone cannot then directly revise a different outcome from that event.

Two lines of work narrow this gap. Espejo et al. (2025) traced cyclone hazards through infrastructure disruption to healthcare access, while Wan et al. (2026) examined mortality and economic-loss mechanisms in a common hazard–exposure–vulnerability framework. Neither tests whether a shared forecasting representation learned from heterogeneous cyclone consequence records lets one reported outcome improve another. Exposure-based deaggregation can create gridded training targets from administrative reports (Kooshki Forooshani et al., 2024; Moss et al., 2026), but its within-area allocation is assumed rather than observed. Stochastic cyclone catalogs expand plausible hazard scenarios (Emanuel et al., 2006, 2008; Loridan and Bruneau, 2025) without supplying observed multi-consequence responses. Together, these studies motivate coordination of impact evidence while keeping the observed administrative scale explicit.

Multi-task learning offers one route: shared features may use common event information, while task-specific heads retain different exposure bases and response mechanisms. Benefits depend on task relatedness, architecture, loss balancing, and control of negative transfer (Caruana, 1997; Baxter, 2000; Ruder, 2017; Sener and Koltun, 2018; Chen et al., 2018; Standley et al., 2020). Disaster applications have mainly used shared text or images for classification (Wang et al., 2020; Alam et al., 2023). Earthquake studies provide closer impact precedents: Patten et al. (2024) tested whether selected observations improved other impacts using separate models, and Loake et al. (2026) jointly modeled several impacts with updates as observations arrived. These precedents make evidence sharing a credible question, but do not establish its reliability for county-level tropical cyclone consequences.

OmniCyRisk tests an integrated response to this evidence problem. It combines a shared spatial representation and consequence-adaptive heads with exposure-consistent county supervision, simulation-assisted initialization, and updates based only on available records. The contribution is evaluated through an architecture-matched single-task comparison of marginal reliability, a diagnostic of observed cross-consequence relationships, and masked tests of whether partial evidence improves consequences, cities, or years excluded from updating. This framing neither assumes that each component is individually new nor that every consequence benefits from sharing. The contribution lies in testing these capabilities together as one evidence-consistent forecasting system, including where coordination helps and where it fails.

More broadly, the experiments outline how unevenly arriving impact reports could inform forecasts for still-unreported outcomes, with separate evaluations of individual accuracy and cross-consequence coherence under fixed evidence. This could support more coherent multi-sector situational awareness, but its value for real-time decisions and its transfer to independent regions require prospective and external validation.

This study addresses three research questions:

- How does coordinated forecasting compare with architecture-matched single-task models for each consequence?
- Does it better preserve observed relationships among simultaneous consequence forecasts?
- How do partial observations affect targets excluded from updating across consequences, cities, and events, and where are gains unreliable?

Simulation-assisted initialization and exposure-consistent aggregation support these tests without treating synthetic responses or latent grid outputs as observed impacts. Fig. 1 summarizes the workflow: isolated experts provide reference forecasts and auxiliary responses for simulated cyclone scenarios, after which OmniCyRisk is refined with historical county-level records. Exposure-consistent aggregation links latent grid-level impact ratios to the county totals used for supervision. We evaluate unseen cyclones under fixed historical evidence, then independently restart from a pre-2019 checkpoint for controlled updates using partially observed records. Each transfer target is excluded from the update.

Section 2 describes the empirical evidence, forecasting formulation, OmniCyRisk architecture, training procedures, experimental designs, and evaluation framework. Section 3 presents the fixed-evidence and partial-evidence results, Section 4 discusses their implications and limitations, and Section 5 concludes the study.

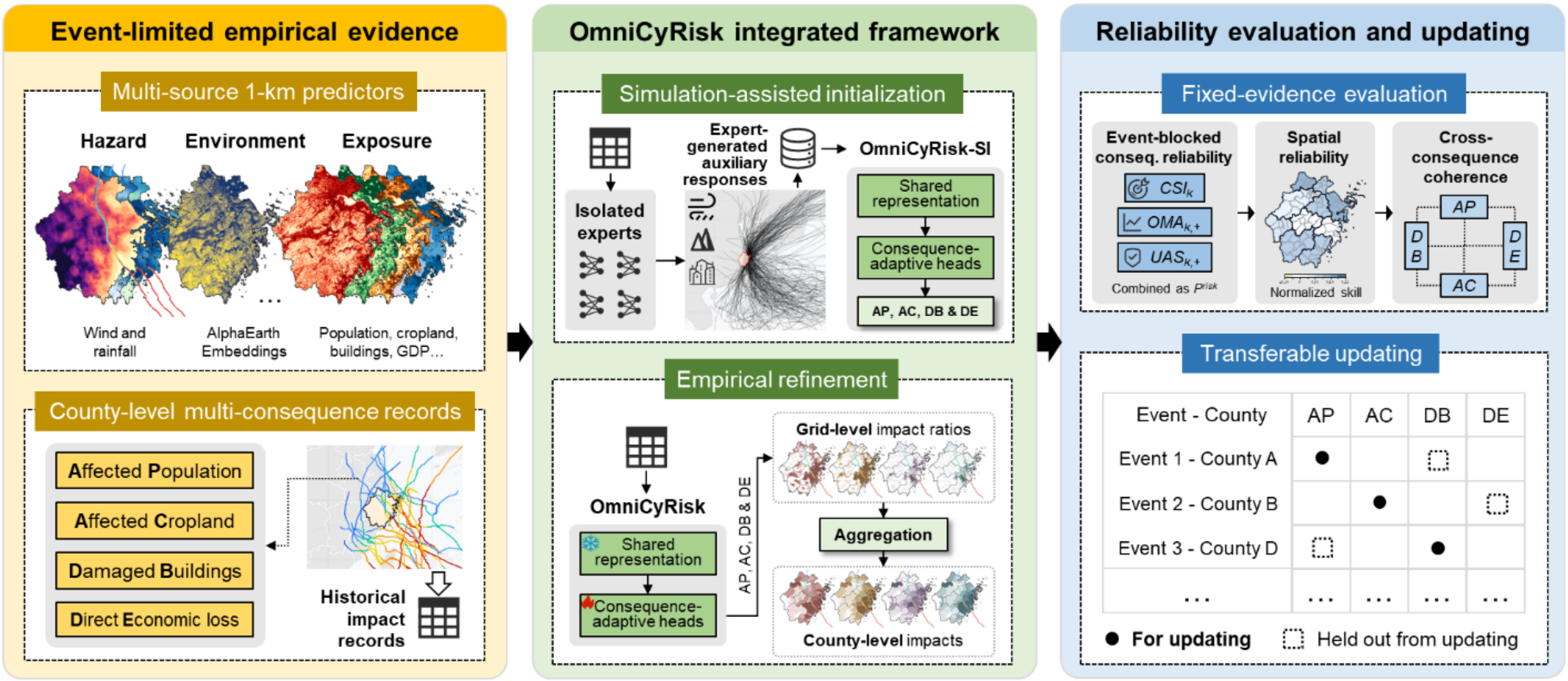

**Fig. 1. Overall workflow of OmniCyRisk**, connecting event-limited multi-consequence evidence, simulation-assisted initialization, exposure-consistent aggregation, reliability-oriented evaluation, and controlled updating under partially observed records.

# 2. Methods

## 2.1 Study region, event archive, and datasets

The study combines county-level records for four consequences from 20 cyclones affecting 90 counties in Zhejiang Province with 71 channels of 1-km hazard, environmental, and exposure predictors. The finalized archive is complete for these four consequences; partial-evidence conditions are generated through controlled masking. Fig. 2 shows the event archive and its temporal and spatial heterogeneity.

### 2.1.1 Study region and historical event archive

Zhejiang Province lies on the southeastern coast of China and is frequently affected by western North Pacific tropical cyclones. Densely populated and highly urbanized coastal plains coexist with major infrastructure and economic concentrations, extensive agricultural areas, and mountainous inland terrain. These contrasts produce substantial spatial variation in hazard exposure, land use, asset concentration, and vulnerability, making Zhejiang a demanding regional testbed for multi-consequence impact forecasting.

The study includes 20 tropical cyclones that affected Zhejiang between 2012 and 2022 and for which province-wide county records were available for all four selected consequences. The spatial domain comprises 90 county-level administrative units. Fig. 2(a) shows the cyclone tracks and intensity evolution, which were obtained from the China Meteorological Administration tropical cyclone best-track archive and used to characterize the historical event archive. Fig. 2(b) shows the province's topographic and population context, based on the DEM product of Abrams et al. (2020) and LandScan Global population data (LandScan Global, 2024). Supplementary Table S1 reports the TC identifier, cyclone name, event dates, analysis-window duration, maximum wind speed, assigned fold, and severity rank for each event.

### 2.1.2 Multi-consequence impact records and event-limited evidence structure

County-level impact records were provided by the Zhejiang Provincial Emergency Management Department. Four consequences are modeled: affected population (AP), affected cropland (AC), damaged buildings (DB), and direct economic loss (DE), representing human, agricultural, built-environment, and aggregate economic impacts, respectively. They describe complementary aspects of the same events but differ substantially in prevalence, scale, and distribution.

The archive contains one value for each of the four consequences in every event–county combination, yielding 1,800 records per consequence and 7,200 consequence-specific observations in total. Zero-valued entries are coded and analyzed as reported zeros rather than missing values. The fixed-evidence experiments therefore use a complete $20 \times 90 \times 4$ event–county–consequence matrix. By contrast, the partial-evidence experiments in Section 2.4.2 deliberately mask selected consequence, year, or city records to test whether available observations benefit targets excluded from updating.

Although each consequence has 1,800 county–event records, those records are nested within only 20 cyclones. Nonzero entries account for 23.8%–40.3% of records, and positive impacts span several orders of magnitude (Table 1). Fig. 2 shows that event severity and within-event county patterns differ across the four outcomes. This combination of event scarcity, sparsity, and heterogeneous scale motivates the consequence-adaptive model and reliability measures in Section 2.5.

### 2.1.3 Gridded hazard, environmental, and exposure predictors

The model-ready predictor dataset contains 71 channels at 1-km resolution: 3 event-specific hazard fields, 64 AlphaEarth Embedding channels, and 4 exposure layers. Together, these predictors represent the hazard forcing, environmental context, and exposed assets involved in the impact-generating process. Table 2 explicitly lists the source, spatial resolution, temporal treatment, and modeling role of each dataset.

Hazard fields were derived from hourly observations collected by approximately 2,500 meteorological stations operated by the Zhejiang Provincial Meteorological Bureau. The station data were interpolated to the common 1-km grid using the same inverse-distance-weighting procedure for every event. Three event summaries were retained: maximum hourly 10-min mean wind speed (MW), maximum hourly rainfall (MR), and maximum 12-h accumulated rainfall (AR).

Environmental context was represented by 64 AlphaEarth Embedding (AEE) channels, which provide a learned representation of land-surface and environmental conditions. Annual AEE layers are available from 2017 onward; consequently, the 2017 layer was used for events from 2012 to 2016, and event-year layers were used from 2017 onward. This approximation and its implications are discussed in Section 4.3.

Exposure was represented by gridded population count (POP), cropland area (CA), building count (BC), and gross domestic product (GDP). These variables serve both as model inputs and as the consequence-specific aggregation bases for AP, AC, DB, and DE, respectively. Source-specific layers were temporally

**Table 1.** Statistical characteristics of county-level multi-consequence impact records.

| Consequence | Unit | Nonzero records (%) | Mean | Maximum | Skewness | Max. $\log_{10}(1+y)$ | Exposure layer |
|---|---|---|---|---|---|---|---|
| Affected population (AP) | persons | 40.3 | 24,203 | 959,193 | **6.40** | 5.98 | Population count |
| Affected cropland (AC) | ha | 31.2 | 1,086 | 34,907 | **4.69** | 4.54 | Cropland area |
| Damaged buildings (DB) | buildings | 23.8 | 150 | 25,650 | **14.16** | 4.41 | Building count |
| Direct economic loss (DE) | 10,000 RMB | 34.4 | 10,812 | 2,065,000 | **20.32** | 6.31 | Gross domestic product (GDP) |

**Note:** Each consequence contains 1,800 county–event records from 20 cyclones and 90 counties. Means and skewness are calculated over all records, including zeros. The county-level records are nested within events and are not statistically independent realizations of 1,800 cyclones.

**Table 2.** Multi-source datasets used for model-ready tropical cyclone impact forecasting.

| Category | Variable | Symbol | Spatial resolution | Temporal treatment | Source | Role in framework |
|---|---|---|---|---|---|---|
| **Hazard** | Maximum hourly 10-min mean wind speed | MW | 1 km | Event-specific, 2012–2022 | Zhejiang Provincial Meteorological Bureau | wind forcing |
| | Maximum hourly rainfall | MR | 1 km | | | Short-duration rainfall intensity |
| | Maximum 12-h accumulated rainfall | AR | 1 km | | | Accumulated rainfall forcing |
| **Environmental** | AlphaEarth Embeddings | AEE | 1 km | 2017 layer used for 2012–2016; annual layers thereafter | Brown et al. (2025) | Land-surface and environmental context |
| **Exposure** | Population count | POP | 1 km | Source-specific layers were matched to study years where available; missing years were estimated by linear interpolation. | LandScan Global (2024) | Input and aggregation base for AP |
| | Cropland area | CA | 1 km | | Gong et al. (2019) | Input and aggregation base for AC |
| | Building count | BC | 1 km | | Che et al. (2024) | Input and aggregation base for DB |
| | Gross domestic product | GDP | 1 km | | Zhao et al. (2017) | Input and aggregation base for DE |

matched to the study events where available, with missing-year values estimated by linear interpolation where necessary. All layers were resampled or aggregated to the common 1-km grid and aligned with the county-boundary masks. All temporal matching, interpolation, and spatial alignment were performed independently of the impact labels.

### 2.1.4 Data limitations and evaluation scope

Three data-related boundaries define the evaluation scope of this study: the scale mismatch between gridded predictors and administrative impact labels, the use of impact records from a single provincial reporting system, and the use of realized rather than forecast hazard fields.

The first boundary is a spatial-scale mismatch: the predictors vary over 1-km cells, whereas the observed impacts are county totals. The models therefore predict grid-level impact ratios, multiply them by the appropriate exposure, and sum within county boundaries before comparing with reported outcomes. This uses fine-resolution environmental information without assigning artificial impact labels to individual cells. The mapped ratios are latent allocations consistent with the county observations; their local accuracy cannot be verified from the present archive.

The impact records are administrative disaster reports from one provincial reporting system in Zhejiang. They provide county-level supervision for the four selected consequences, but their definitions and verification procedures may reflect local reporting practice. Event-blocked cross-validation tests new cyclones within this setting. The cross-city experiment tests transfer of new observations to cities represented during base-model development; neither protocol measures transfer to an independent province or country. That test requires an external archive with comparable observed consequence definitions, spatial units, and event periods, together with exposure and hazard inputs aligned to those records. Global exposure coverage alone cannot establish forecast skill in a new reporting system.

The historical experiments also use hazard fields derived from meteorological observations. They therefore evaluate impact prediction conditional on realized wind and rainfall, not the complete operational chain from meteorological forecasting to impact estimation. Forecast meteorological fields with the same input structure could be supplied in deployment, but the propagation of meteorological forecast uncertainty lies outside the scope of this study.

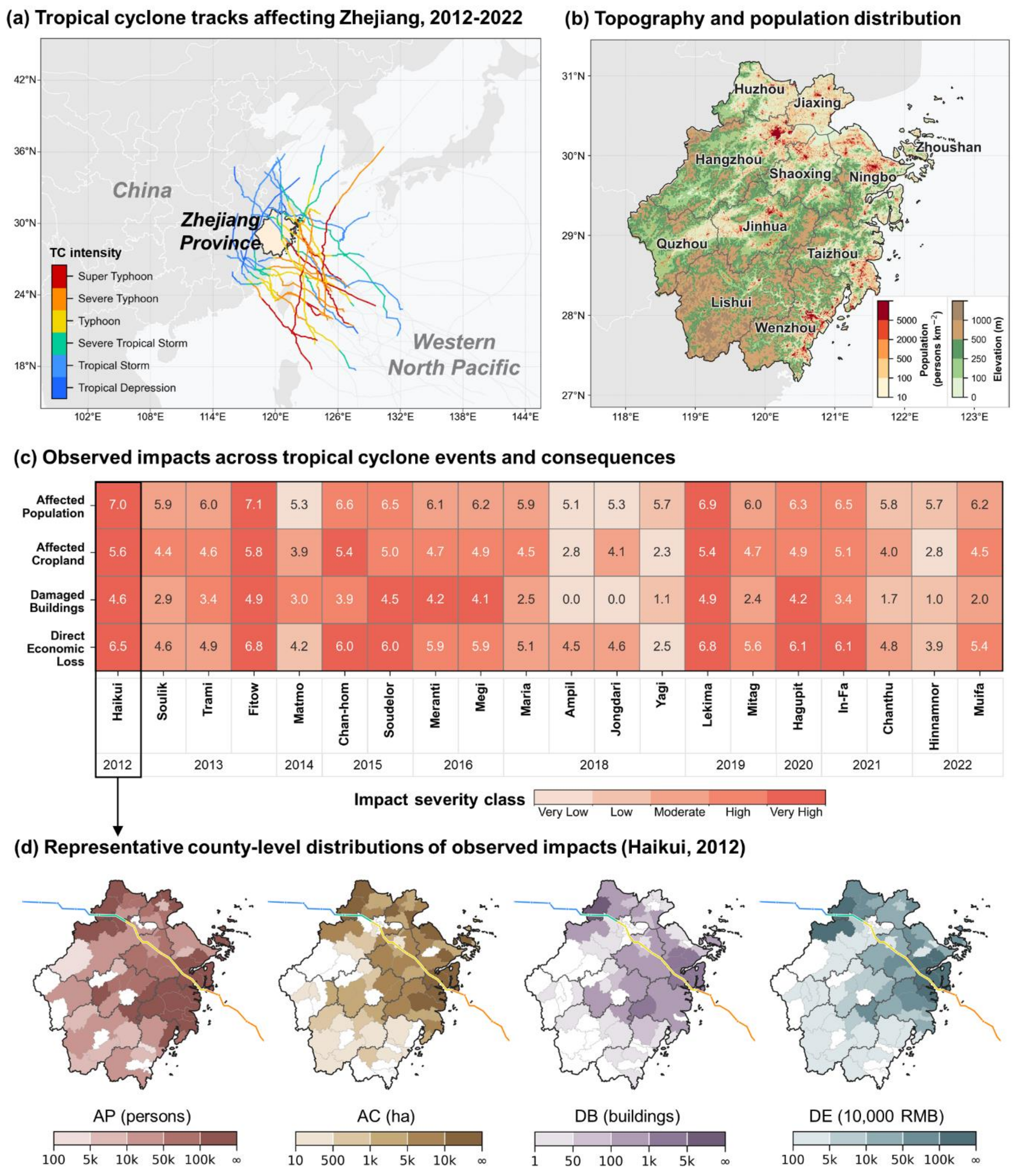

**Fig. 2. Study region, historical event archive, and structure of multi-consequence impact evidence.** (a) Tracks and intensities of the 20 tropical cyclones affecting Zhejiang Province during 2012–2022, based on the China Meteorological Administration tropical cyclone best-track archive. (b) Provincial topography and population distribution. (c) Event-level severity of affected population (AP), affected cropland (AC), damaged buildings (DB), and direct economic loss (DE), derived from county-level disaster impact records provided by the Zhejiang Provincial Emergency Management Department. Values were calculated by summing county-level records within Zhejiang for each event; cell values report log10(1+province-wide total impact), and colors indicate consequence-specific ordinal severity classes. (d) County-level distributions of the four observed consequences for Typhoon Haikui (2012), illustrating related but non-identical spatial impact patterns.

These data boundaries determine both the observation scale and the scope of the tests. Section 2.2 defines the forecasting task and the competing models; Section 2.3 then explains how they are trained.

## 2.2 Forecasting formulation and models

The formulation separates three questions that recur throughout the paper: what is observed, how a model turns gridded predictors into the observed county totals, and what differs between independent and coordinated forecasting. The same exposure-consistent output rule is used by every comparison model. This common rule lets the later experiments focus on learning and information sharing rather than on different definitions of the prediction target.

### 2.2.1 Task definition and exposure-consistent aggregation

Let $\boldsymbol{\mathcal{E}} = \{1, 2, \ldots, N\}$ denote the set of historical tropical cyclone events. For each event $e \in \boldsymbol{\mathcal{E}}$, the spatial domain is represented at two levels. The fine-resolution modeling domain consists of grid cells $g \in \boldsymbol{\mathcal{G}}$, whereas observed disaster-impact records are reported for administrative units $c \in \boldsymbol{\mathcal{C}}$, such as counties. Let $\boldsymbol{\mathcal{G}}(c) \subseteq \boldsymbol{\mathcal{G}}$ denote the set of grid cells contained in county c.

For event e, the gridded predictor field is written as

$$\mathbf{X}_e = \{\mathbf{x}_{e,g}: g \in \mathcal{G}\} \tag{1}$$

where $\mathbf{x}_{e,g} \in \mathbb{R}^p$ is a p-dimensional feature vector describing hazard intensity, environmental context, exposure, and vulnerability-related conditions at grid cell g. In this study, the predictors include tropical cyclone wind and rainfall variables, environmental descriptors, and exposure layers at 1-km resolution.

Let $\boldsymbol{\mathcal{K}} = \{1, 2, \ldots, K\}$ denote the set of disaster-impact consequences. For consequence $k \in \boldsymbol{\mathcal{K}}$, the observed impact of event e in county c is denoted by $y_{e,c}^{(k)}$, and the corresponding county-level impact field is

$$\mathbf{Y}_e^{(k)} = \{y_{e,c}^{(k)}: c \in \mathcal{C}\} \tag{2}$$

The forecasting objective is to estimate all consequence fields from the same event-level predictor field:

$$\mathbf{X}_e \mapsto \left\{\hat{\mathbf{Y}}_{e,F}^{(k)}\right\}_{k \in \boldsymbol{\mathcal{K}}} \tag{3}$$

where $F$ denotes the forecasting system.

Because predictors are gridded while supervision is available as county totals, both the isolated and coordinated systems use the same fine-to-coarse output formulation. For forecasting system F ∈ {iso, Omni}, let $\hat{r}_{e,g,\mathrm{F}}^{(k)}$ denote the predicted grid-level impact ratio for consequence k. The corresponding county-level estimate is obtained through exposure-consistent aggregation:

$$\hat{y}_{e,\mathrm{c},\mathrm{F}}^{(k)} = \sum_{\mathrm{g} \in \mathcal{G}(\mathrm{c})} \mathrm{a}_{e,\mathrm{g}}^{(k)} \hat{r}_{e,\mathrm{g},\mathrm{F}}^{(k)}; \quad \mathrm{c} \in \mathcal{C} \tag{4}$$

where $\mathrm{a}_{e,\mathrm{g}}^{(k)}$ is the consequence-specific exposure quantity at grid cell g. Population count, cropland area, building count, and gridded gross domestic product serve as the exposure bases for affected population, affected cropland, damaged buildings, and direct economic loss, respectively. The predicted county-level field is therefore

$$\hat{\mathbf{Y}}_{e,\mathrm{F}}^{(k)} = \{\hat{y}_{e,\mathrm{c},\mathrm{F}}^{(k)} : \mathrm{c} \in \mathcal{C}\} \tag{5}$$

The four outputs have distinct units and response distributions but share aspects of the same hazard, exposure, and vulnerability context.

The county records are nested within only 20

cyclones, so their nominal count does not provide 1,800 independent hazard realizations per consequence. During a partial-evidence update, an unobserved outcome is masked rather than coded as zero; only available records enter the updating loss.

Both isolated and coordinated models multiply predicted grid-level impact ratios by the relevant exposure and sum within county boundaries (Eqs. (4) and (5)). Training loss and quantitative evaluation use the resulting county estimates, the scale at which impacts were reported. The grid-level fields may help explain how the model distributes an event spatially, but multiple local allocations could be compatible with the same county total. They should therefore be read as latent model outputs, not as independently verified 1-km forecasts.

The aggregation rule is common to every model in this study. What differs is how the models produce the grid-level ratios that enter it: the next subsection contrasts independent and coordinated prediction pathways.

### 2.2.2 Isolated and coordinated forecasting formulations

A conventional forecasting system develops one independent model for each consequence. For consequence $k$, the isolated prediction is

$$\{\hat{r}_{e,\mathrm{g,iso}}^{(k)}: \mathrm{g} \in \mathcal{G}\} = f_k(\mathbf{X}_e; \boldsymbol{\theta}_k^{\mathrm{iso}}), \qquad k \in \mathcal{K} \quad (6)$$

where $f_k(\cdot)$ is a consequence-specific model with parameters $\boldsymbol{\theta}_k^{\mathrm{iso}}$. This formulation permits specialization to the distribution and response mechanisms of each consequence. However, the representation for consequence $k$ is learned exclusively from its own records; observations of other impacts from the same events cannot directly inform it.

OmniCyRisk instead constructs a coordinated multi-consequence system. The predictor field is first transformed into a shared disaster representation:

$$\mathbf{H}_e = \phi(\mathbf{X}_e; \boldsymbol{\theta}_s), \quad (7)$$

where $\phi(\cdot)$ is the shared feature extractor with $\boldsymbol{\theta}_s$. Consequence-adaptive prediction functions then map the shared representation to distinct grid-level impact-ratio fields:

$$\{\hat{r}_{e,\mathrm{g,Omni}}^{(k)}: \mathrm{g} \in \mathcal{G}\} = g_k(\mathbf{H}_e; \boldsymbol{\theta}_k), \quad k \in \mathcal{K} \quad (8)$$

In Eq. (8), each consequence has its own prediction function and adaptive parameters, but all four functions receive the shared representation in Eq. (7). This is where evidence from different consequences can inform a common forecasting system. Section 2.2.3 details the coordinated architecture, and Section 2.2.4 describes the models used for comparison.

### 2.2.3 OmniCyRisk architecture

OmniCyRisk separates shared spatial representation learning from consequence-specific impact prediction. For each tropical cyclone, the model receives the 71-channel, 1-km input tensor described in Section 2.1: 3 event-specific hazard variables, 64 AlphaEarth embedding channels, and 4 exposure variables representing population, cropland, buildings, and gross domestic product. As shown in Fig. 3(a), patch embedding, a hierarchical Swin Transformer encoder, and a U-Net-style decoder transform these predictors into a shared high-resolution representation. Four consequence-adaptive mixture-of-experts (MoE) heads then generate separate grid-level impact-ratio fields for affected population (AP), affected cropland (AC), damaged buildings (DB), and direct economic loss (DE). The predicted ratios are subsequently transformed into county-level impact forecasts through exposure-consistent aggregation.

Exposure serves two distinct functions. As part of the input tensor, it helps the model represent local asset concentration and hazard–exposure interactions. As the consequence-specific aggregation base, it converts each predicted dimensionless ratio into an absolute county-level impact. These roles use the same exposure information but enter the framework at different stages.

The shared backbone has three components (Fig. 3(b1–b3)). Patch embedding partitions the input grid into local patches and projects them into feature tokens with positional information. Successive Swin Transformer stages combine windowed and shifted-window self-attention with patch merging to capture multiscale interactions among hazard gradients, exposure concentrations, terrain, and land-surface conditions. The U-Net-style decoder then restores spatial resolution through upsampling and skip connections, producing the shared feature map supplied to the prediction heads.

Each consequence-adaptive MoE head combines shared and consequence-specific expert branches through input-dependent softmax gating (Fig. 3(b4)). The gated features are processed by a consequence-specific fusion and output layer, allowing the four consequences to use common spatial information while retaining distinct response functions. A final 1×1 convolution and sigmoid activation produce a grid-level impact ratio bounded between 0 and 1. Architectural dimensions, attention heads, window sizes, expert configurations, normalization, and dropout settings are reported in Supplementary Table

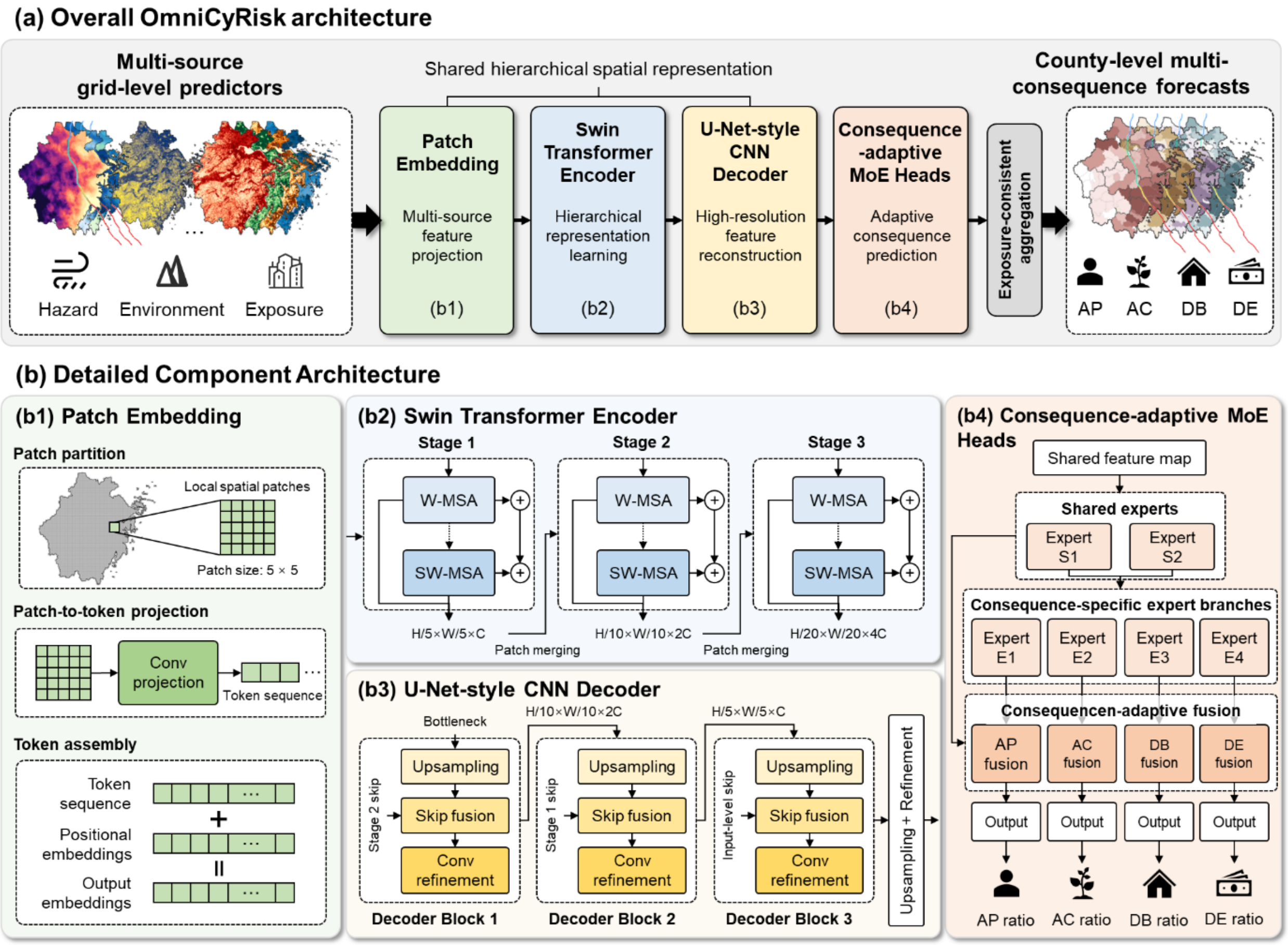

**Fig. 3. OmniCyRisk architecture for multi-consequence tropical cyclone impact forecasting.** (a) Overall mapping from 1-km hazard, environmental, and exposure predictors to county-level forecasts of AP, AC, DB, and DE. A shared hierarchical backbone learns the spatial representation, and four consequence-adaptive MoE heads generate distinct grid-level impact-ratio fields that are converted to county totals through exposure-consistent aggregation. (b) Component architecture: (b1) patch partition and embedding; (b2) hierarchical Swin Transformer encoding; (b3) U-Net-style feature reconstruction; and (b4) consequence-adaptive MoE heads combining shared and consequence-specific expert branches through learned softmax gating.

S2a.

For consequence $k$, the head produces the ratio field defined in Eq. (8). Exposure-consistent aggregation in Eqs. (4) and (5) then converts that field into county-level impact estimates using population, cropland, buildings, or GDP as the respective aggregation base. OmniCyRisk and the isolated models use the same predictors, output definition, and grid-to-county aggregation, so their forecasts are evaluated under a common observation model.

The complete OmniCyRisk model contains 3,210,284 trainable parameters. After empirical refinement, it performs stand-alone multi-consequence inference; the isolated experts used for comparison and initialization are not required at prediction time.

### 2.2.4 Comparison models and capacity

Four isolated convolutional neural networks were developed for AP, AC, DB, and DE and are denoted ISO-AP, ISO-AC, ISO-DB, and ISO-DE, respectively. Their architecture follows the tropical cyclone impact model of Lin and Wang (2024), adapted to the common input, output, and aggregation formulation used here. Each expert maps the same 71-channel predictor tensor to one grid-level impact-ratio field, applies a 1×1 output convolution and sigmoid activation, and aggregates the predicted ratios to county impacts using Eqs. (4) and (5). Detailed architectural configurations and training settings of the isolated experts are provided in Supplementary Table S2b.

Each ISO expert has 43,393 trainable parameters, compared with 3,210,284 for OmniCyRisk. The ISO comparison therefore evaluates complete forecasting strategies rather than matched model capacity. Both use the same predictors, targets, output definition, aggregation rule, and event-blocked evaluation.

The ISO experts supply held-out reference forecasts and county-level auxiliary responses for the simulated cyclone library. Those responses transfer relationships learned from historical records into OmniCyRisk initialization without adding empirical events. Experts and auxiliary targets were constructed within each training fold; held-out events were excluded from fitting and simulation-assisted initialization (Section 2.4.1).

The experiments use these models for distinct purposes. The ISO experts supply a lightweight end-to-end reference and simulated auxiliary targets. OmniCyRisk-RI and OmniCyRisk-SI differ only in whether they receive simulation-assisted initialization; their comparison tests the effect of that starting point under the same empirical refinement. Omni-STL independently trains each consequence with matched architecture, initialization, optimization, aggregation, and evaluation settings, providing the principal single-task comparison for coordinated learning. This matches each prediction pathway, although the four independent models do not have the same combined parameter budget as one joint model. Unless stated otherwise, OmniCyRisk denotes SI.

## 2.3 Simulation-assisted initialization and training

The higher-capacity OmniCyRisk model must learn from only 20 independent cyclones. Simulation-assisted initialization supplies a starting representation before refinement with observed multi-consequence records (Fig. 4).

The simulation library contains 240 event-level wind and rainfall realizations over Zhejiang Province on the same 1-km grid as the historical predictors. Each realization is combined with the corresponding environmental and exposure layers. The simulated hazard scenarios are generated independently of the impact records, as described in Supplementary Methods S1. Their auxiliary consequence responses are subsequently produced by the ISO experts trained on the real training partition.

Training proceeds in three stages. **First,** ISO-AP, ISO-AC, ISO-DB, and ISO-DE are fitted independently to their observed training records. **Second,** these experts generate AP, AC, DB, and DE responses for the 240 simulated scenarios. The resulting responses are model-generated auxiliary targets, not additional empirical observations. **Third,** all OmniCyRisk components are initialized by optimizing against these auxiliary targets and are then refined using observed multi-consequence training records. During empirical refinement, the patch embedding, Swin Transformer encoder, and U-Net decoder are frozen; only the consequence-adaptive MoE heads and their loss-weighting parameters are updated. Approximately 1% of the model parameters are therefore modified during this stage.

This sequence broadens the hazard configurations encountered during initialization while reserving observed impact records for empirical refinement, validation-based model selection, and held-out evaluation. The 240 simulated scenarios should consequently be interpreted as structured initialization support rather than as 240 additional observed cyclones.

Training uses an asymmetric loss on log-transformed county impacts because positive reports span several orders of magnitude. Underestimation receives five times the weight of overestimation of the same transformed magnitude, reflecting the cost of missing a substantial impact while retaining a continuous error signal. Each ISO expert minimizes its own consequence loss. OmniCyRisk combines the four losses with learned uncertainty weights and a softplus regularizer that prevents a consequence loss from being made arbitrarily small through its weight (Kendall et al., 2018). The exact loss and weighting equations remain available in Supplementary Methods S2.1, while the main text keeps the objective and its purpose together with the training sequence.

The ISO experts optimize all model parameters against their respective $L_k$. During simulation-assisted initialization, all OmniCyRisk parameters are optimized against $L_{\mathrm{Omni}}$ using the expert-generated auxiliary targets. During empirical refinement, the shared backbone remains frozen and only the consequence-adaptive heads and $s_k$ parameters are optimized against observed impacts. Exact architecture settings, optimization hyperparameters, stopping criteria, and model-selection rules are reported in Supplementary Table S2.

The completed training sequence supplies a fixed model for two different evaluations. Event-blocked tests ask how it predicts cyclones excluded from fitting; independent retrospective updates ask whether new partial records improve targets excluded from those updates. The protocols below keep these questions separate.

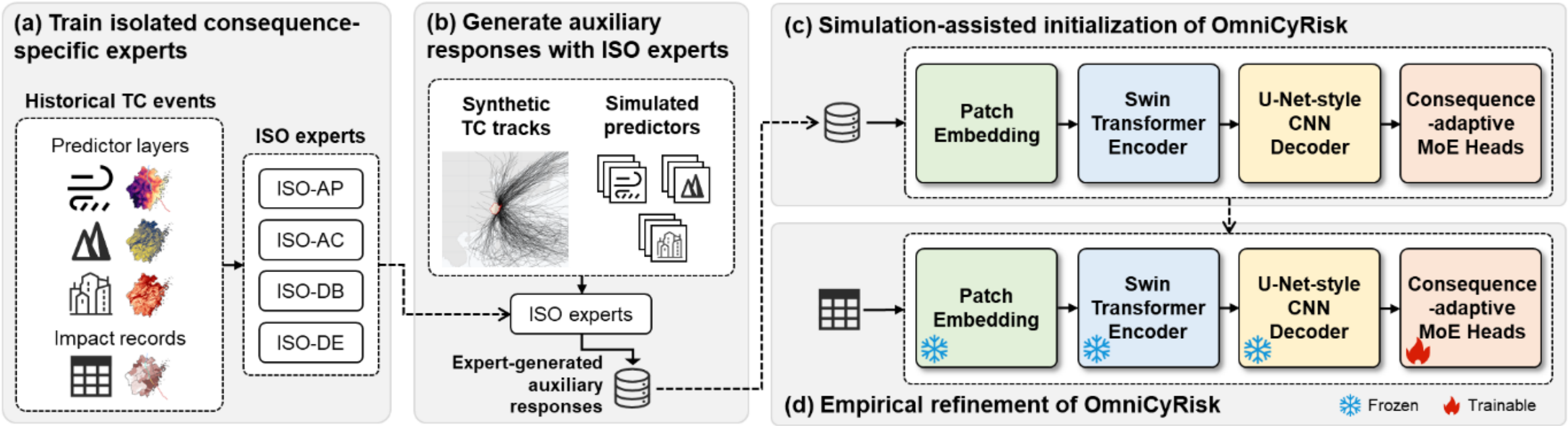

**Fig. 4. Fold-specific simulation-assisted initialization and empirical refinement of OmniCyRisk.** (a) ISO-AP, ISO-AC, ISO-DB, and ISO-DE are trained separately using historical impact records from the training partition. (b) The fitted experts generate auxiliary consequence responses for 240 simulated predictor scenarios. (c) All OmniCyRisk components are optimized using the simulated predictors and expert-generated responses. (d) Observed training records refine the consequence-adaptive MoE heads while the patch embedding, Swin Transformer encoder, and U-Net decoder remain frozen. Validation records are used only for model selection, and test impact records are excluded from all model-development stages.

### 2.4 Experimental designs

#### 2.4.1 Fixed-evidence event-blocked evaluation

All 20 cyclones were evaluated in five complete event-blocked folds, so counties from one cyclone never appeared in both training and test partitions. Four events were held out for testing in each iteration. Fold allocation was stratified by mean severity across the four consequences to avoid highly imbalanced partitions. The remaining 16 events were used for training and validation, while test events were excluded from model fitting, preprocessing estimation, parameter selection, and empirical refinement. Event assignments and fold characteristics are reported in Supplementary Table S1.

Within each partition, the ISO experts were fitted to the real training records and applied to the fixed simulation library to generate fold-specific auxiliary responses. OmniCyRisk was initialized using the simulated predictors and these responses, then refined using the observed multi-consequence training records. Validation records were used only for model selection. All data-dependent preprocessing parameters were estimated from the training partition and applied unchanged to validation and test data.

#### 2.4.2 Partial-evidence updating experiments and controls

Observation masks restrict retrospective updates to available consequence–event–county records. A masked outcome contributes neither a target label nor a zero to the updating objective.

For consequence $k$, event $e$, and county $c$, record availability is represented by

$$m_{e,c}^{(k)} \in \{0,1\} \tag{9}$$

where $m_{e,c}^{(k)}=1$ indicates that the record is observed and included in the updating objective, whereas $m_{e,c}^{(k)}=0$ excludes it.

Let $\Omega_{\text{new}}$ denote the event-county indices considered in an update, and let

$$N_k^{\text{obs}} = \sum_{(e,c)\in\Omega_{\text{new}}} m_{e,c}^{(k)}$$

denote the number of observed records for consequence $k$. The masked consequence-level updating loss is

$$\mathcal{L}_{k,\text{upd}} = \frac{\sum_{(e,c)\in\Omega_{\text{new}}} m_{e,c}^{(k)}\,\ell_{e,c}^{(k)}}{\max\left(1, N_k^{\text{obs}}\right)} \tag{10}$$

where $\ell_{e,c}^{(k)}$ is the loss for one observed record. The consequence-level losses are combined using the multi-consequence objective defined in Section 2.3. Normalization by $N_k^{\text{obs}}$ prevents a consequence from dominating an update solely because it has more reported records.

Let $\mathcal{D}_{\text{new}}^{\text{obs}}$ denote the observed portion of the newly available evidence. The updated forecasting system is written as

$$F_{\text{new}} = \mathcal{U}\left(F_{\text{base}}; \mathcal{D}_{\text{new}}^{\text{obs}}\right) \tag{11}$$

where $F_{\text{base}}$ and $F_{\text{new}}$ denote the systems before and after updating.

In OmniCyRisk, available records can modify components shared across the coordinated prediction heads and thereby influence excluded targets. In the implemented protocol, the hierarchical backbone remains frozen and updating is restricted to the consequence-adaptive MoE heads; the experiments therefore evaluate transfer through the coordinated head architecture rather than through backbone relearning.

Section 3.3 evaluates withheld consequences, cities, and years. These independent retrospective experiments test transfer within an established Zhejiang system, not zero-shot regional generalization or chronological online deployment.

A common base model, $F_{\text{base}}$, was established using tropical cyclones occurring up to 2018 (Fig. 5a). The consequence-specific experts used for simulation-assisted initialization were trained exclusively on pre-2019 historical records, ensuring that post-2018 observations did not contribute to the auxiliary responses. OmniCyRisk was initialized using expert-generated responses for the simulated cyclone scenarios and then refined using the available pre-2019 empirical records. The resulting checkpoint represents the established forecasting system before post-2018 evidence becomes available.

Each updating experiment began independently from the same $F_{\text{base}}$ checkpoint (Fig. 5b–d). Following the empirical updating procedure in Section 2.3, the shared backbone remained frozen and updating was restricted to the consequence-adaptive mixture-of-experts (MoE) heads. The corresponding updating configuration and hyperparameter settings are summarized in Supplementary Table S2a. Available records were used only for updating; all target records used for evaluation were excluded from the updating loss, normalization, model selection, and hyperparameter determination. Pre- and post-update predictions were evaluated on identical held-out targets. Their difference estimates the net effect of updating with partial evidence on held-out targets, without fitting the model to the evaluation records. Each protocol withholds one target while updating from the complementary post-2018 records: one of four consequences, one of 11 cities, or one target

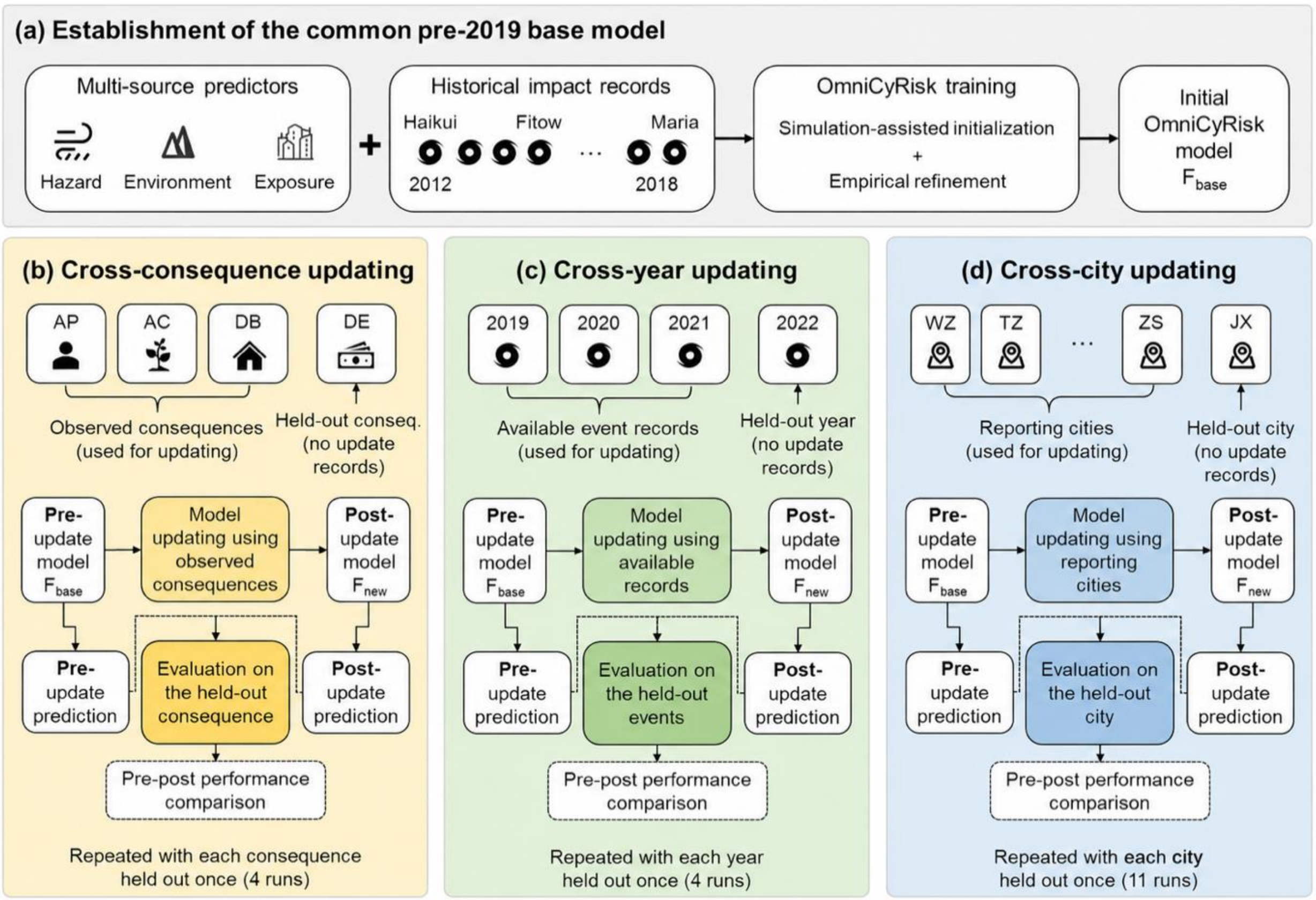

**Fig. 5. Controlled updating framework for OmniCyRisk under partially observed impact evidence.** (a) Establishment of the common pre-2019 base model $F_{\text{base}}$. (b) Cross-consequence updating, using three observed consequences while withholding the fourth. (c) Cross-year updating, using records from other years while withholding one target year. (d) Cross-city updating, using records from other cities while withholding each of the 11 cities in turn. Every experiment restarts independently from $F_{\text{base}}$, and the held-out target is excluded from the updating loss, normalization, model selection, and hyperparameter determination.

year (Fig. 5).

For cross-consequence updating, the withheld consequence supplies no observed label to the update. Its pre-update predictions serve as fixed self-consistency targets in the update loss to prevent drift, alongside the loss on newly available observations from the other three consequences. Any change in the withheld forecast therefore comes indirectly through adaptable shared components within the MoE heads. In the cross-city and cross-year protocols, all observed labels from the target city or year are withheld.

Two additional checks clarify the updating mechanism and its cost. A frozen-shared-expert control kept the shared expert branches within the MoE heads fixed and updated only the consequence-specific components. Differences between this control and full updating quantify the additional contribution of adapting the shared head components; the common Swin backbone remained frozen in both cases. Retention was assessed on a fixed pre-2019 reference subset that was reserved before development of $F_{\text{base}}$ and excluded from model fitting and selection.

## 2.5 Evaluation measures and statistical uncertainty

### 2.5.1 Consequence reliability and cross-consequence coherence

We assess three distinct failure modes: missed affected counties, inaccurate magnitudes where an impact occurred, and systematic underestimation. The corresponding components are the critical success index (CSI), positive-impact order-of-magnitude agreement (OMA+), and underestimation avoidance (UAS+). The exact log transformation, classification threshold, and component formulae are given in Supplementary Methods S2.2 (Eqs. (S4)–(S7)).

The distinction is needed because most county–event records are zero, while positive impacts are highly skewed. A single average regression error could reward a model that predicts zeros well yet misses affected counties, or allow a few large impacts to dominate the assessment. The three components instead show where a forecast succeeds or fails before they are combined. Continuous-error diagnostics remain useful, but they answer a different question from this reliability score.

The impact-classification threshold is selected using validation records within each fold and then held fixed for its test events. OMA+ is calculated on observed positive impacts and uses a one-order-of-magnitude tolerance. That deliberately broad tolerance acknowledges the range and reporting uncertainty of the administrative data; it is a severity-agreement diagnostic rather than exact-value accuracy. UAS+ decreases as substantial underprediction accumulates in transformed space. Keeping zeros out of the positive-impact agreement component prevents their greater frequency from overwhelming the severity assessment. These measures describe predictive failure modes; they are not direct measurements of evacuation or resource allocation outcomes.

$$P_k^{\text{risk}}(F;\mathcal{T}) = \left[\text{CSI}_k\ \text{OMA}_{k,+}\ \text{UAS}_{k,+}\right]^{1/3} \qquad (12)$$

The geometric mean limits compensation among components. We average consequence scores for the macro score and report normalized skill relative to the stated comparator as the fraction of remaining score headroom recovered. The exact macro and normalized-skill formulae are in Supplementary Methods S2.2 (Eqs. (S8)–(S10)). A change in this diagnostic score has no direct numerical conversion to decision benefit.

Consequence-level performance does not by itself establish whether simultaneous forecasts form a coherent impact profile. Let $\rho_{k\ell}$ and $\hat{\rho}_{k\ell,F}$ denote the observed and predicted Spearman correlations between consequences $k$ and $\ell$, calculated from the same log-transformed held-out samples. Cross-consequence coherence error is defined as

$$E_{\text{coh}}(F) = \frac{2}{K(K-1)} \sum_{k<\ell} \left|\rho_{k\ell} - \hat{\rho}_{k\ell,F}\right|. \qquad (13)$$

Lower coherence error means that the six predicted pairwise rank correlations are closer, on average, to those observed in the same held-out county–event records. It asks whether four simultaneous forecasts preserve the ordering relationships among impacts, a property that consequence-level scores alone cannot show. Pairwise rank agreement remains a limited diagnostic: it neither validates the complete joint impact distribution nor replaces verification of each consequence.

### 2.5.2 Update skill and statistical uncertainty

Let $h$ index a target excluded from the updating loss, such as a consequence, year, or region. Its normalized update skill is

$$S_h^{\mathrm{upd}} = \frac{P_{h,\mathrm{post}}^{\mathrm{risk}} - P_{h,\mathrm{pre}}^{\mathrm{risk}}}{1 - P_{h,\mathrm{pre}}^{\mathrm{risk}}} \tag{14}$$

For a held-out consequence, $P_h^{\mathrm{risk}}$ is its consequence-level score. For a held-out city, the target score is the macro-average across consequences within that city. For cross-year updating, we calculate a macro score for each held-out cyclone and average these scores equally across the seven cyclones, while withholding all events in the target year from updating. A positive value indicates that partial evidence improves a target that contributed no observations to the updating loss.

We report target-equal macro update skill and pooled update skill separately. The former gives each withheld consequence, city, or cyclone-level macro score (for cross-year updating) one equal contribution to the summary. The latter recomputes the comparison after combining the relevant held-out observations or consequence scores; targets with different record counts and baseline headroom can therefore affect it differently. The two summaries answer related but distinct questions and should not be interchanged. Their exact definitions are in Supplementary Methods S2.3.

Across protocols, positive normalized update skill indicates improvement on a target excluded from updating. Tables 5 and 6 report target, macro, and pooled confidence intervals from the paired event–county bootstrap. In Table 7, individual cyclone rows are descriptive point estimates; the seven-cyclone pooled result carries the confidence interval.

Confidence intervals for both fixed-evidence comparisons and update-skill estimates use 1,000 paired event–county bootstrap resamples. The same held-out indices are applied to both forecasts in each resample, so the intervals describe paired record-level variability in this archive. They quantify the uncertainty in estimated performance differences, rather than predictive uncertainty for individual forecasts. Although fixed-evidence evaluation uses event-blocked folds, the bootstrap does not fully capture dependence among counties affected by the same cyclone. With only 20 independent cyclones, and seven targets in the cross-year test, these intervals should not be interpreted as full uncertainty for future events.

## 3. Results

### 3.1 Fixed-evidence predictive reliability

We first assess predictive reliability under fixed historical evidence. The analysis compares OmniCyRisk with architecture-matched Omni-STL, examines event-level agreement for the 20 cyclones, and evaluates the contribution of simulation-assisted initialization.

#### 3.1.1 Matched single-task comparison

The principal fixed-evidence comparison is with Omni-STL, whose per-consequence architecture matches OmniCyRisk while its four models learn independently, without cross-consequence sharing. Both systems use the same initialization, optimization, aggregation, and event-blocked evaluation settings. Supplementary Table S4a reports fold results, and Supplementary Table S4b reports pooled component metrics. This architecture-matched comparison examines the role of cross-consequence sharing, although the combined parameter budgets cannot be matched.

OmniCyRisk achieved higher $P_k^{\mathrm{risk}}$ values in 14 of the 20 consequence–fold comparisons (Supplementary Table S4a), showing that the advantage occurred in individual event-blocked folds as well as in the pooled analysis. Pooling all out-of-fold predictions showed that OmniCyRisk achieved comparable predictive reliability to Omni-STL with a modest overall improvement (Table 3). The macro-average reliability score increased from 0.675 for Omni-STL to 0.685 for OmniCyRisk, corresponding to a normalized skill of 3.1% [95% CI: 0.7%–5.3%]. The gains were consequence-dependent: AP and AC showed improvements, DE remained largely unchanged, whereas DB exhibited a negative change. These heterogeneous responses indicate that cross-consequence sharing does not provide a universal accuracy advantage, and its contribution varies among consequences with different response characteristics.

Component-level differences further clarify the reliability changes (Table 4). At the macro level, $\mathrm{CSI}_k$ increased by 1.5 percentage points [0.5, 2.5], while $\mathrm{UAS}_{k,+}$ increased by 1.1 percentage points [0.7, 1.5]. The change in $\Delta\mathrm{OMA}_{k,+}$ was small and statistically uncertain (0.2 percentage points [−1.2, 1.6]), indicating that the net differences in this matched comparison lay mainly in impact identification and underestimation avoidance, while positive-impact magnitude agreement remained broadly comparable; these components do not isolate a unique mechanism.

The matched comparison establishes competitive marginal reliability, with a modest aggregate gain and worse damaged-building performance. For context, OmniCyRisk also exceeded the lightweight ISO experts in 17 of 20 consequence–fold comparisons (Supplementary Table S3). That end-to-end comparison is capacity-imbalanced and does not isolate the contribution of sharing. The matched result supplies the accuracy foundation for the system-level tests.

Table 3. Event-blocked out-of-fold reliability of OmniCyRisk and Omni-STL.

| Consequence | Omni-STL $P_k^{\mathrm{risk}}$ | OmniCyRisk $P_k^{\mathrm{risk}}$ | Normalized skill vs. Omni-STL (%) [95% CI] |
|---|---|---|---|
| **AP** | 0.707 | 0.740 | **11.5** [7.5, 15.6] |
| **AC** | 0.659 | 0.688 | **8.4** [4.7, 12.3] |
| **DB** | 0.658 | 0.631 | -7.8 [-12.3, -3.6] |
| **DE** | 0.678 | 0.682 | **1.2** [-3.4, 5.7] |
| **Macro average** | **0.675** | **0.685** | **3.1** [0.7, 5.3] |

**Note:** Omni-STL is the architecture-matched single-task counterpart of OmniCyRisk, trained independently for each consequence without cross-task sharing. Normalized skill is calculated relative to Omni-STL.

**Table 4.** Component-wise reliability differences between OmniCyRisk and Omni-STL.

| Consequence | $\Delta CSI_k$ (pp) | $\Delta OMA_{k,+}$(pp) | $\Delta UAS_{k,+}$(pp) |
|---|---|---|---|
| **AP** | 4.5 | 3.4 | 1.3 |
| **AC** | 5.4 | 0.5 | 1.3 |
| **DB** | -3.6 | -4.4 | 1.6 |
| **DE** | -0.3 | 1.3 | 0.2 |
| **Macro average [95% CI]** | **1.5 [0.5, 2.5]** | **0.2 [-1.2, 1.6]** | **1.1 [0.7, 1.5]** |

**Note:** Values are paired differences between OmniCyRisk and Omni-STL, multiplied by 100 and expressed in percentage points (pp), not relative percentage changes. Positive values favor OmniCyRisk and negative values favor Omni-STL. Estimates are calculated from the same event-blocked out-of-fold predictions used for the composite risk score. Confidence intervals are obtained using the paired event-county bootstrap. Corresponding model-specific component values and paired event-county bootstrap confidence intervals are reported in Supplementary Table S4b.

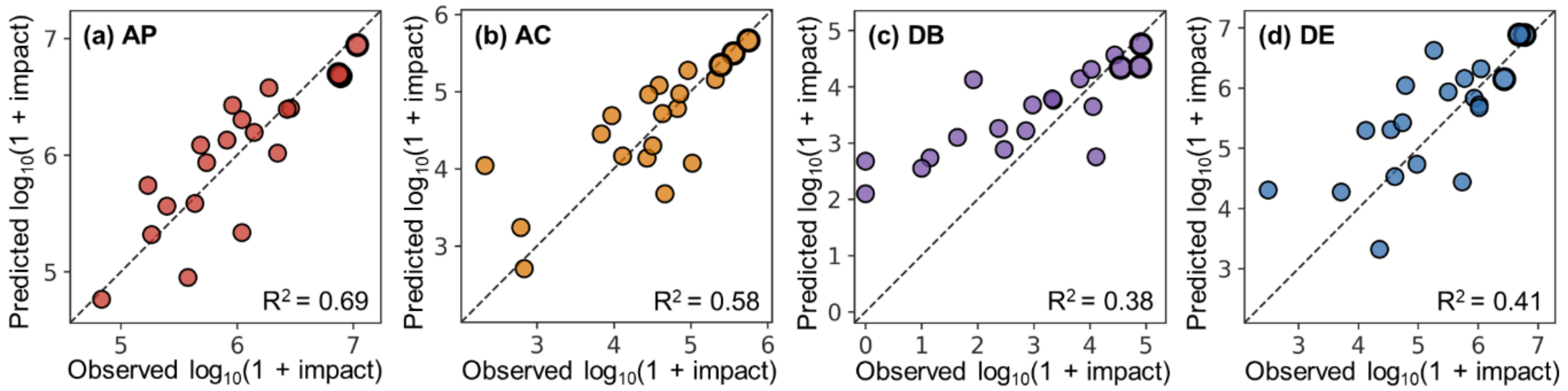

**Fig. 6. Observed versus out-of-fold predicted province-wide impacts for 20 cyclones**: (a) AP, (b) AC, (c) DB, and (d) DE. Axes show $\log_{10}(1 + \text{impact})$; $R^2$ is calculated on this scale. Outlined points mark Fitow (2013), Lekima (2019), and Haikui (2012), the three highest-ranked events by observed multi-consequence severity. Dashed lines show perfect agreement.

### 3.1.2 Event-level verification

The observed multi-consequence severity ranking identifies Fitow (2013), Lekima (2019), and Haikui (2012) as the three highest-impact events in this archive (Fig. 2). Their CMA best-track maximum 2-min sustained wind speeds within the analysis windows were 45, 62, and 45 m $s^{-1}$, respectively (Supplementary Table S1). This ranking does not follow maximum wind speed alone; the realized wind and rainfall footprint, its overlap with exposed assets, and local vulnerability also influence reported impacts. Storm radius was not analyzed in this study, so we do not infer a quantitative size–damage relation.

Fig. 6 compares observed and out-of-fold predicted province-wide totals for all 20 cyclones. On the $\log_{10}(1 + \text{impact})$ scale, observed–predicted $R^2$ is 0.69 for affected population, 0.58 for affected cropland, 0.38 for damaged buildings, and 0.41 for direct economic loss. Larger-impact events generally receive higher predictions, while the lower DB and DE agreement makes clear that event severity is not estimated equally well for all consequences. These are held-out event predictions, not fits to the displayed cyclones.

### 3.1.3 Contribution of simulation-assisted initialization

OmniCyRisk-SI and OmniCyRisk-RI share architecture, empirical records, and the same frozen-backbone refinement protocol; they differ in their starting parameters (Section 2.2.4). The RI comparison is therefore an initialization ablation under this protocol, not a comparison with end-to-end training from random parameters.

Across all five folds, SI began empirical refinement at lower validation loss and followed a shorter, more stable optimization path than RI (Fig. 7a). Several RI folds showed larger training–validation divergence. The optimization advantage generally carried through to held-out performance. OmniCyRisk-SI achieved higher test-fold $P_k^{\text{risk}}$ in 19 of the 20 consequence–fold comparisons and in all five macro-average comparisons (Fig. 7(b)). The single exception occurred for DB in Fold 4. Under the tested frozen-backbone refinement protocol, the simulated starting point improved optimization and final reliability in nearly every event partition; this does not test end-to-end training from random parameters.

The simulated scenarios broaden the wind and rainfall configurations encountered before empirical refinement. Their consequence responses were generated by experts trained on the historical fold, so they organize relationships already learned from observed events rather than add 240 new observed cyclones. Their errors can also be inherited by the initialized model. The SI–RI comparison isolates the practical effect of this starting point under the same later refinement and held-out test.

### 3.2 System-level coherence and spatial variation

We next examine two system-level properties beyond marginal reliability: how closely the forecasts preserve observed cross-consequence relationships and how comparative performance varies across cities and county groups.

#### 3.2.1 Cross-consequence coherence

We compared pairwise Spearman rank correlations among the four observed consequences with those in pooled out-of-fold Omni-STL and OmniCyRisk predictions (Section 2.5.1). OmniCyRisk matched the observed correlation more closely for five of six pairs (Fig. 8); AP–DB was the exception. Mean absolute pairwise correlation error was 0.064 for Omni-STL and 0.026 for OmniCyRisk, a descriptive reduction of 59.4%. Uncertainty for this coherence difference was not estimated.

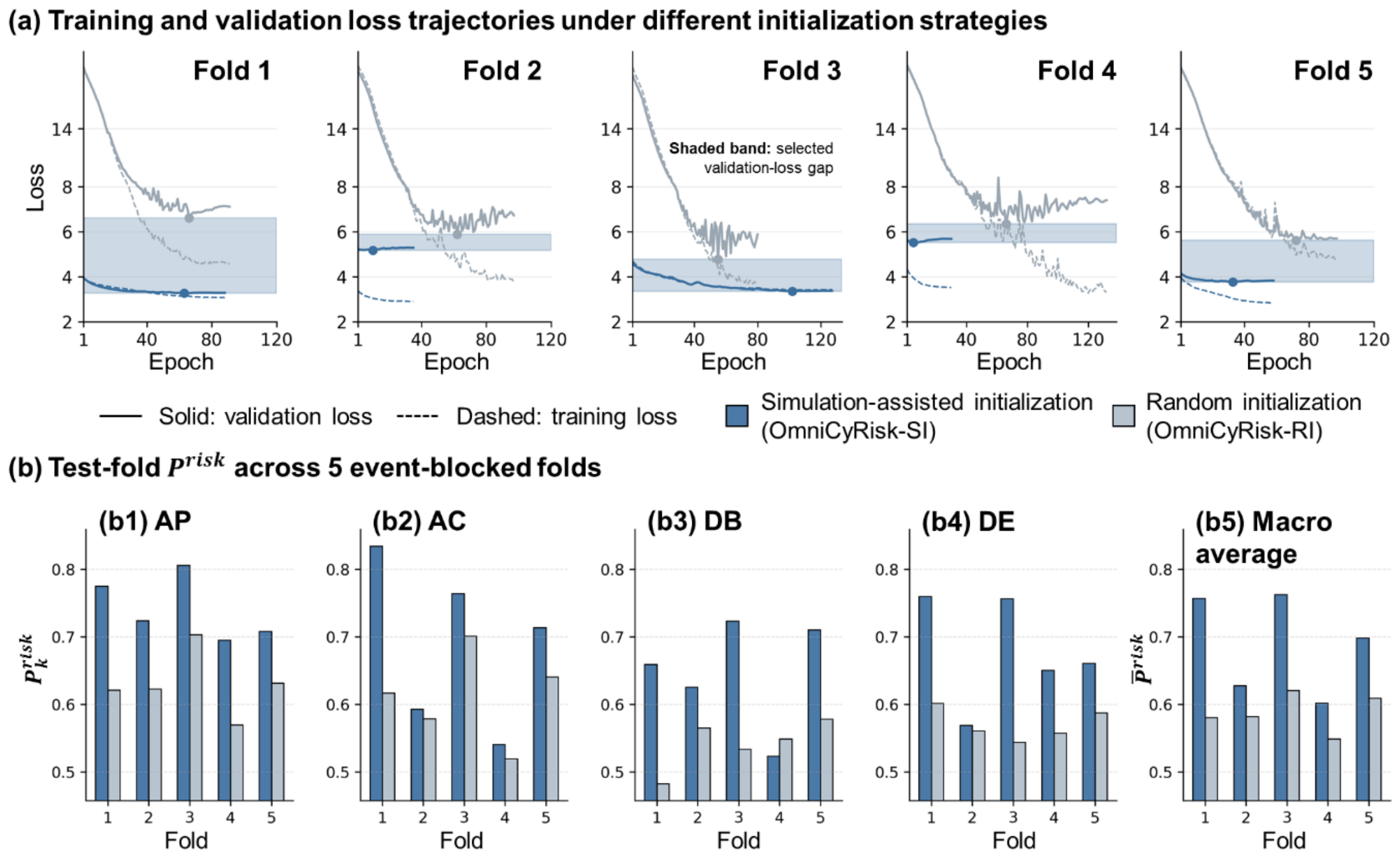

**Fig. 7. Effect of simulation-assisted initialization on optimization and held-out reliability.** (a) Training and validation loss trajectories for OmniCyRisk-SI and OmniCyRisk-RI across the five event-blocked folds. Solid and dashed curves denote validation and training losses, respectively; blue and gray denote simulation-assisted and random initialization. The shaded horizontal band in each fold spans the selected validation-loss values of the two initialization strategies, highlighting their gap. (b) Test-fold $P_k^{\text{risk}}$ for AP, AC, DB, DE, and the macro average. OmniCyRisk-SI exceeds OmniCyRisk-RI in 19 of 20 consequence–fold comparisons and all five macro-average comparisons.

This coherence result complements the modest, heterogeneous marginal reliability comparison. An impact vector can have individually plausible components yet reproduce the observed relationships among human, agricultural, building, and economic losses poorly. The lower pairwise error is therefore informative about the collection of forecasts, while the AP–DB exception and the lack of a coherence confidence interval limit its interpretation. Pairwise rank correlations do not validate the full joint impact distribution or identify a unique causal effect of sharing.

### 3.2.2 Spatial heterogeneity

We examined pooled out-of-fold skill relative to Omni-STL across all 11 cities and county groups defined by coast, terrain, and population density (Fig. 9). Nine cities had positive macro skill, whereas Jinhua and Taizhou had slightly negative values. Consequence-specific composite-score differences also varied in direction. The modest province-wide gain therefore describes an average over heterogeneous places and outcomes, rather than an improvement at every city or for every consequence.

Both categories in each coast, terrain, and population-density stratification had positive macro skill, but the gains varied (Fig. 9). Supplementary Methods S3 gives the group definitions and sample counts. As all groups come from Zhejiang, this analysis describes within-province variation, not skill in an independent region. Together with the two negative city results, it qualifies the matched accuracy finding without altering the separate coherence result.

The fixed-evidence tests therefore establish a modest, heterogeneous accuracy result, a substantial descriptive reduction in pairwise coherence error, and a near-uniform benefit from simulation-assisted initialization. Those tests use no new impact reports after training. We next examine what changes when partial reports become available while the evaluation target remains unobserved.

## 3.3 Transfer under partially observed evidence

Each transfer experiment restarts from the same pre-2019 model and excludes its target from updating and selection. The three protocols test transfer across

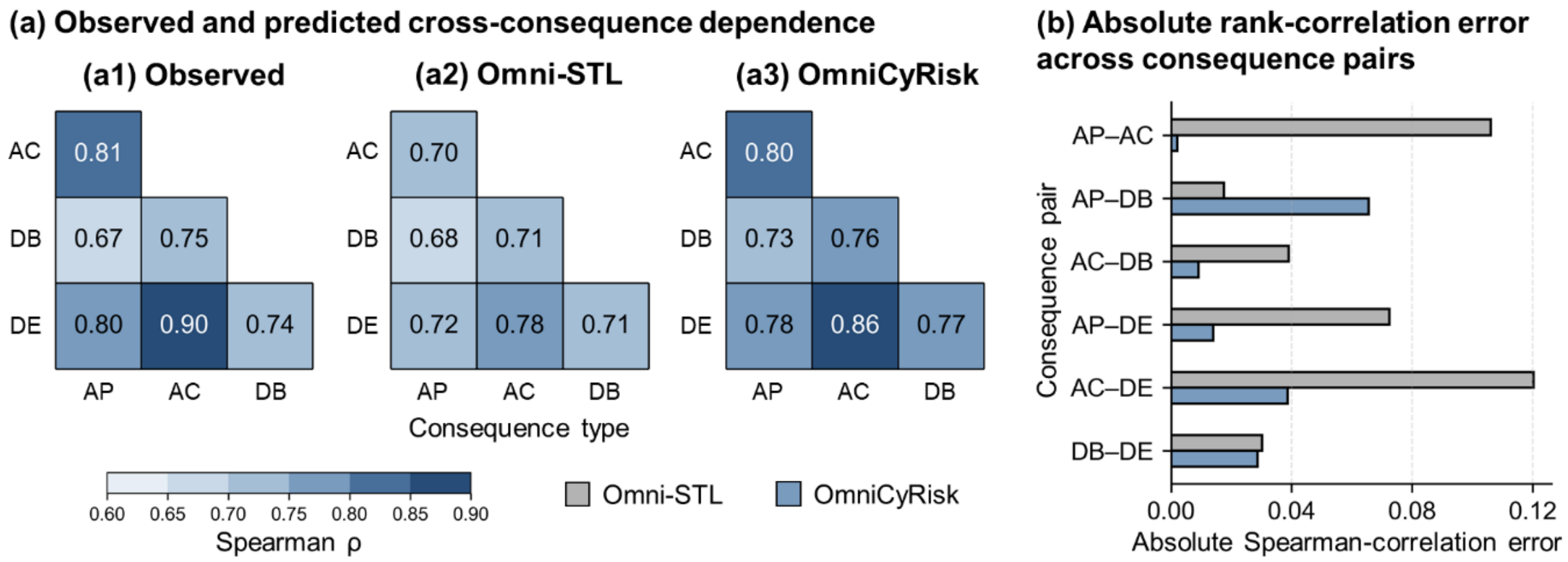

**Fig. 8. Cross-consequence coherence of pooled out-of-fold forecasts.** (a) Pairwise Spearman rank correlations among AP, AC, DB, and DE for (a1) observed impacts, (a2) Omni-STL predictions, and (a3) OmniCyRisk predictions. (b) Absolute rank-correlation errors relative to the observed dependence structure. OmniCyRisk reduced the error for five of the six consequence pairs and decreased the mean coherence error from 0.064 to 0.026; AP–DB was the only exception.

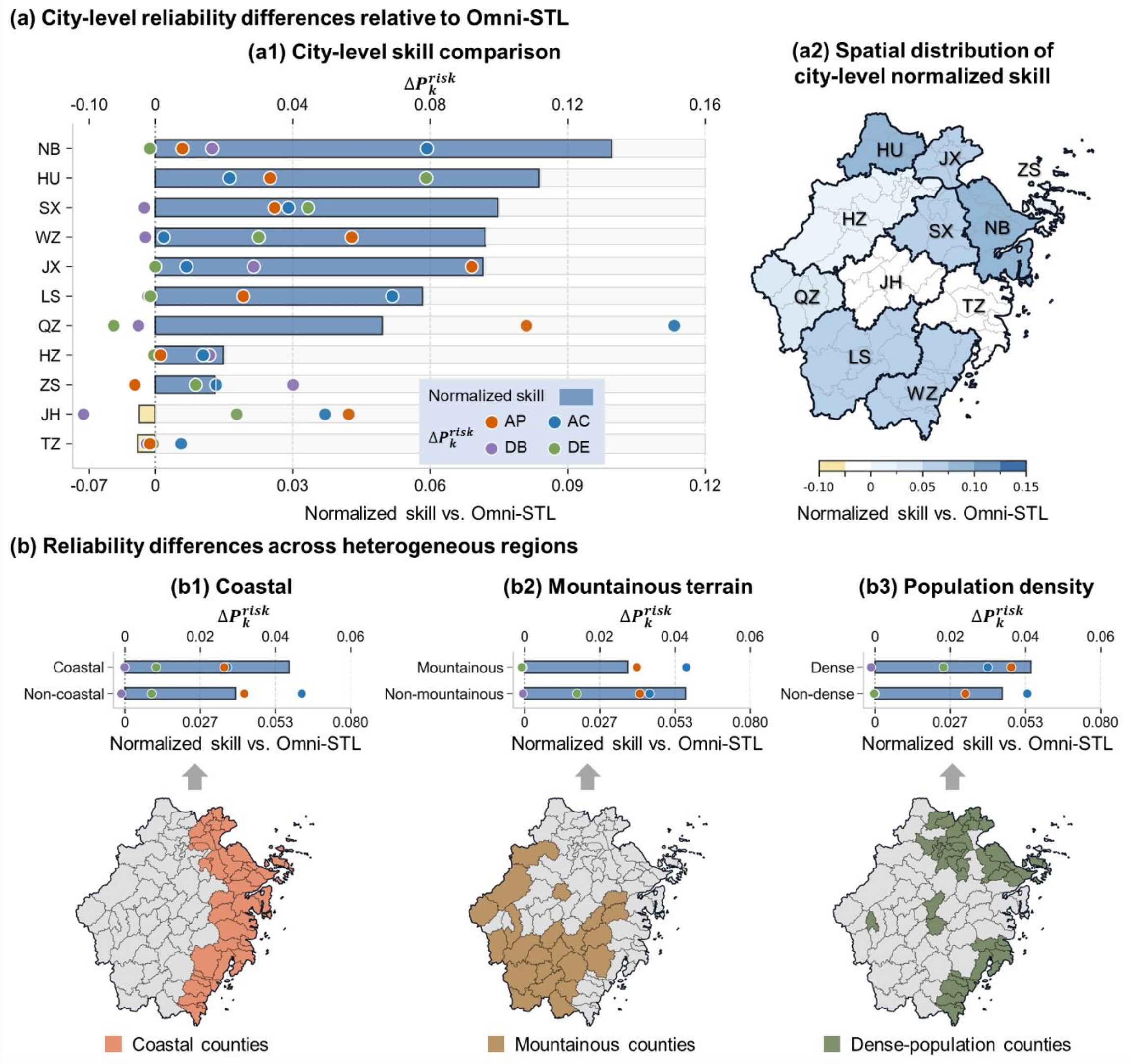

**Fig. 9. Spatial variation in the OmniCyRisk–Omni-STL reliability comparison.** (a) City-level macro normalized skill and consequence-specific composite-score differences, with a map of all 11 cities. (b) Corresponding comparisons for coastal location, mountainous terrain, and population-density county groups. Bars show macro skill; points show consequence-specific differences. Positive values favor OmniCyRisk; negative values favor Omni-STL.

consequences, cities, and years rather than a single chronological update sequence.

### 3.3.1 Cross-consequence transfer to unavailable consequences

Under the cross-consequence protocol in Section 2.4.2, each consequence was withheld in turn while the other three supplied update records.

Updating improved all four unavailable consequences (Table 5). Reliability increased from 0.565 to 0.675 for AP, 0.581 to 0.740 for AC, 0.380 to 0.482 for DB, and 0.500 to 0.659 for DE. Target-specific update skills ranged from 16.6% to 37.8%, with positive record-level bootstrap intervals. Macro-average skill was 27.9%, and pooled skill was 26.9% [21.7%, 28.7%]. Under this retrospective protocol, observations of related consequences improved predictions for outcomes excluded from updating; the score gains have not been translated into decision-level benefit.

Residuals decreased for all four withheld consequences (Fig. 10a). The Lekima examples show

**Table 5.** Cross-consequence updating performance on unavailable consequences

| Held-out consequence | Pre-update $P_k^{risk}$ | Post-update $P_k^{risk}$ | Normalized update skill (%) [95% CI] |
|---|---|---|---|
| **AP** | 0.565 | 0.675 | **25.4** [18.0, 33.7] |
| **AC** | 0.581 | 0.740 | **37.8** [27.2, 41.3] |
| **DB** | 0.380 | 0.482 | **16.6** [7.1, 19.8] |
| **DE** | 0.500 | 0.659 | **31.7** [24.7, 38.7] |
| **Macro average** | **0.506** | **0.639** | **27.9** [22.9, 30.0] |
| **Pooled** | - | - | **26.9** [21.7, 28.7] |

**Note:** Macro skill gives equal weight to the four withheld consequences; pooled skill combines their pre- and post-update reliability before normalization (Supplementary Methods S2.3, Eqs. (S11) and (S12)). Positive skill is the fraction of remaining headroom recovered. Confidence intervals are shown for target, macro, and pooled estimates.

changes in latent grid ratios and county forecasts (Fig. 10b,c); only county-level observations support the quantitative evaluation.

Against the frozen-shared-expert control, full updating added pooled skill of 26.3% [20.1%, 28.8%] (Supplementary Table S5a). Retention was mixed: on the pre-2019 reference set, AC and DE scores fell by 0.048 and 0.032, respectively, with intervals below zero; the macro change was −0.014 with an interval spanning zero. These prior-performance losses qualify the gains on withheld consequences.

### 3.3.2 Cross-city transfer to non-reporting cities

Each of the 11 Zhejiang cities was withheld in turn while other cities supplied update records. The withheld city contributed no labels, normalization information, or selection data. This assesses transfer of new evidence within previously represented regions, not zero-shot prediction in unseen cities.

All 11 held-out cities had positive point estimates of update skill, although the confidence intervals for several individual cities included zero (Supplementary Table S7). Across all cities, the macro-average reliability score increased from 0.557 to 0.696, corresponding to macro update skill of 29.1% [20.7%, 34.5%]. The pooled score increased from 0.617 to 0.712, yielding pooled update skill of 24.9% [18.7%, 30.5%]. Coastal and inland macro skills were 18.5% [10.9%, 23.8%] and 37.9% [25.5%, 48.0%], respectively (Table 6). These within-Zhejiang results show transfer to non-reporting cities across both groups, with substantial variation among individual locations.

For Wenzhou during Lekima, records from other cities shifted predicted AP, AC, and DE county totals toward their observed values (Fig. 11). Wenzhou itself supplied no records to the update. The mapped grid ratios show how the model redistributed its estimates within the city; the county comparisons provide the quantitative evidence because no independently observed grid impacts are available. This case illustrates the direction of one update, while Table 6 and Supplementary Table S7 establish the across-city pattern and its uncertainty.

Full updating exceeded the frozen-shared-expert control by pooled additional skill of 20.8% [14.6%,

**Table 6.** Summary of cross-city updating performance across the 11 held-out prefecture-level cities.

| City group | Pre-update $\overline{P}^{risk}$ | Post-update $\overline{P}^{risk}$ | Normalized update skill (%) [95% CI] |
|---|---|---|---|
| **Coastal macro average** | **0.654** | **0.730** | **18.5** [10.9, 23.8] |
| **Inland macro average** | **0.476** | **0.668** | **37.9** [25.5, 48.0] |
| **Macro average** | **0.557** | **0.696** | **29.1** [20.7, 34.5] |
| **Pooled** | **0.617** | **0.712** | **24.9** [18.7, 30.5] |

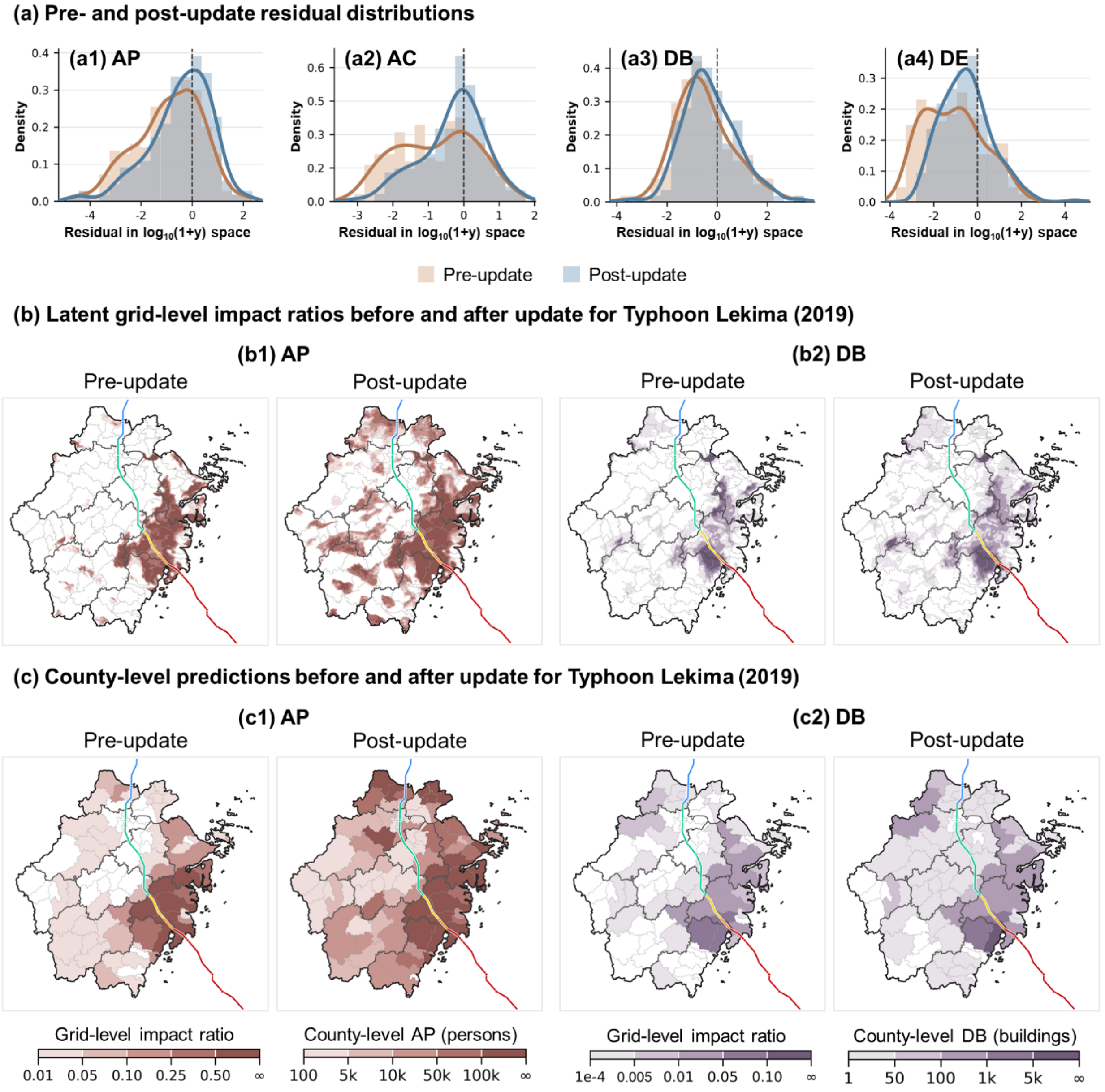

**Fig. 10. Transfer to unavailable consequences through cross-consequence updating.** (a) Pre- and post-update residual distributions for AP, AC, DB, and DE when each is withheld in turn. (b) Illustrative latent grid-level impact-ratio fields for AP and DB during Typhoon Lekima (2019). (c) Corresponding exposure-aggregated county-level predictions. Grid-level fields are latent model outputs; quantitative evaluation is conducted at the county level.

26.6%] (Supplementary Table S5c), consistent with a contribution from adaptable shared head branches. On the pre-2019 reference set, pooled performance fell by 0.037 [−0.045, −0.030]. Thus, gains for non-reporting cities came with measurable loss on prior cases; individual-city intervals and this retention cost both matter when judging an update (Supplementary Table S7).

### 3.3.3 Retrospective cross-year transfer

For each target year from 2019 to 2022, all events in that year were withheld; other post-2018 years supplied update records. The protocol tests retrospective cross-event transfer without target-year labels.

Cross-year updating produced heterogeneous event-level responses (Table 7). Macro-average update skill was 5.8%, and pooled reliability increased from 0.487 to 0.524, equivalent to pooled update skill of 7.2% [95% CI: 1.9%, 10.6%]. Five of seven events had positive point estimates, from 3.8% to 18.9%, while Hinnamnor and Muifa had small negative estimates. The aggregate gain therefore does not imply benefit for every new event. The available seven-event analysis suggests that source–target relevance may matter, but it cannot establish a prospective criterion for choosing records from other years.

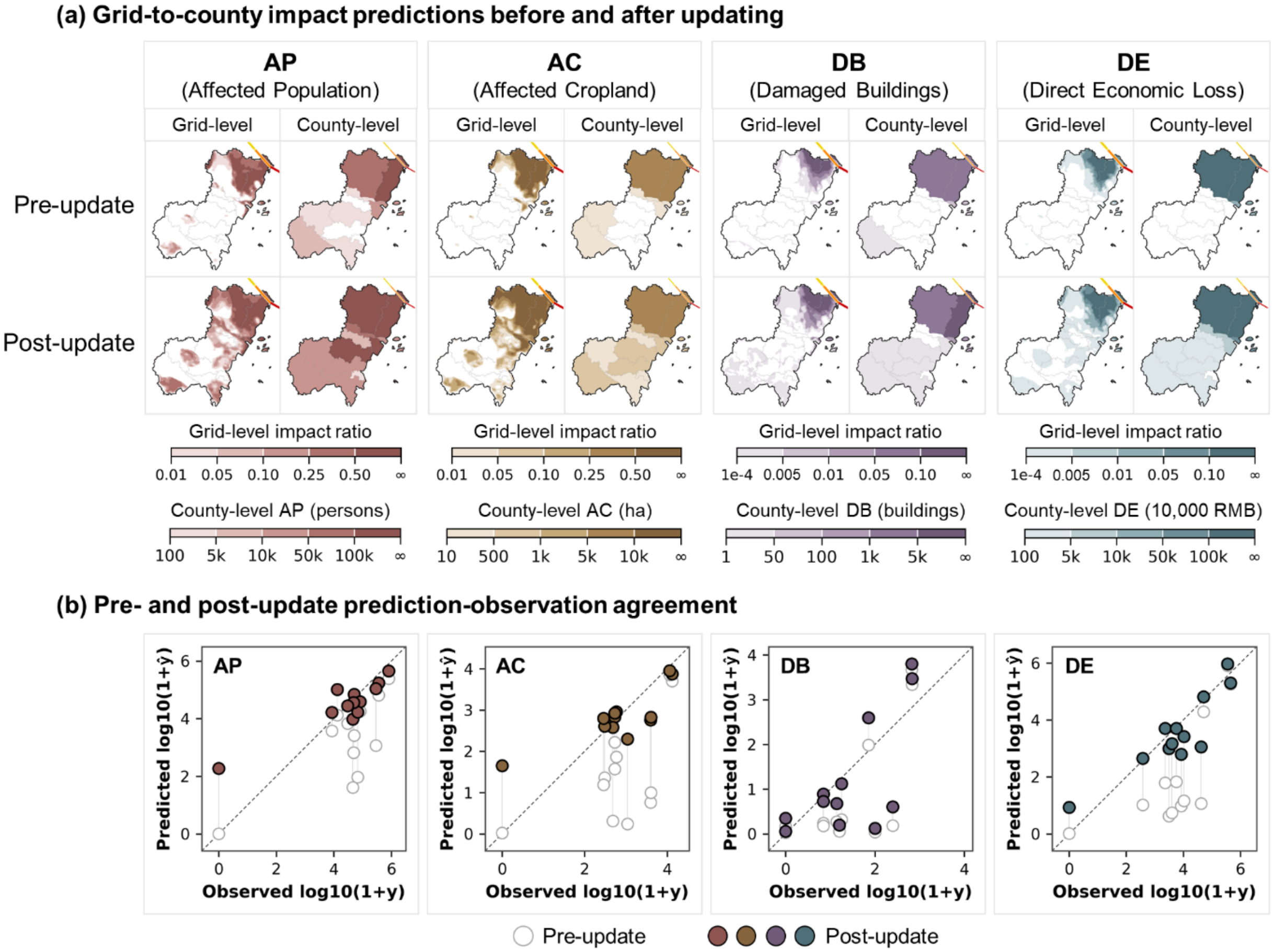

**Fig. 11. Cross-city updating for a non-reporting region: Wenzhou during Typhoon Lekima (2019).** (a) Pre- and post-update grid-level impact-ratio fields and exposure-aggregated county-level predictions for AP, AC, DB, and DE. (b) County-level prediction–observation comparisons before and after updating. Wenzhou supplied no records to the update. Grid-level fields are latent model outputs; quantitative evaluation is conducted at the county level.

Lekima illustrates a favorable cross-year case: absolute residuals decreased in 58/90 counties for AP, 59/90 for AC, 48/90 for DB, and 58/90 for DE (Fig. 12). The event-wise table gives the full, heterogeneous result.

A post-hoc diagnostic associated greater impact-profile distance with lower update skill (Spearman $\rho = -0.54$, $n = 7$), while hazard-descriptor distance showed little association ($\rho = 0.11$; Supplementary Table S6). Muifa remained a negative-transfer case despite a relatively close impact-profile analogue. Because the diagnostic uses realized target outcomes unavailable when screening a new event, it cannot establish a transfer mechanism or a prospective source-selection rule. Such a rule would require descriptors available before target impacts are known and validation on events not used to design it.

Full updating added pooled skill of 3.6% [−1.4%, 5.6%] over the frozen-shared-expert control (Supplementary Table S5b), an interval spanning zero. Pre-2019 reference performance was essentially unchanged. Before operational use, candidate updates require testing on representative historical events and monitoring of retained performance; the present seven-event result does not validate an acceptance rule.

**Table 7.** Cross-year updating performance on held-out tropical cyclone events

| Held-out event | Pre-update $\overline{P}^{risk}$ | Post-update $\overline{P}^{risk}$ | Normalized update skill (%) [95% CI] |
|---|---|---|---|
| **Lekima (2019)** | 0.446 | 0.503 | **10.4** |
| **Mitag (2019)** | 0.779 | 0.792 | **5.9** |
| **Hagupit (2020)** | 0.696 | 0.754 | **18.9** |
| **In-Fa (2021)** | 0.268 | 0.312 | **6.0** |
| **Chanthu (2021)** | 0.529 | 0.547 | **3.8** |
| **Hinnamnor (2022)** | 0.237 | 0.217 | -2.6 |
| **Muifa (2022)** | 0.646 | 0.640 | -1.7 |
| **Macro average** | **0.514** | **0.538** | **5.8** |
| **Pooled** | **0.487** | **0.524** | **7.2** [1.9, 10.6] |

**Note:** Individual event rows and the macro average are reported as descriptive point estimates. The pooled 95% confidence interval was obtained using paired event–county bootstrap resampling within the seven post-2018 target cyclones.

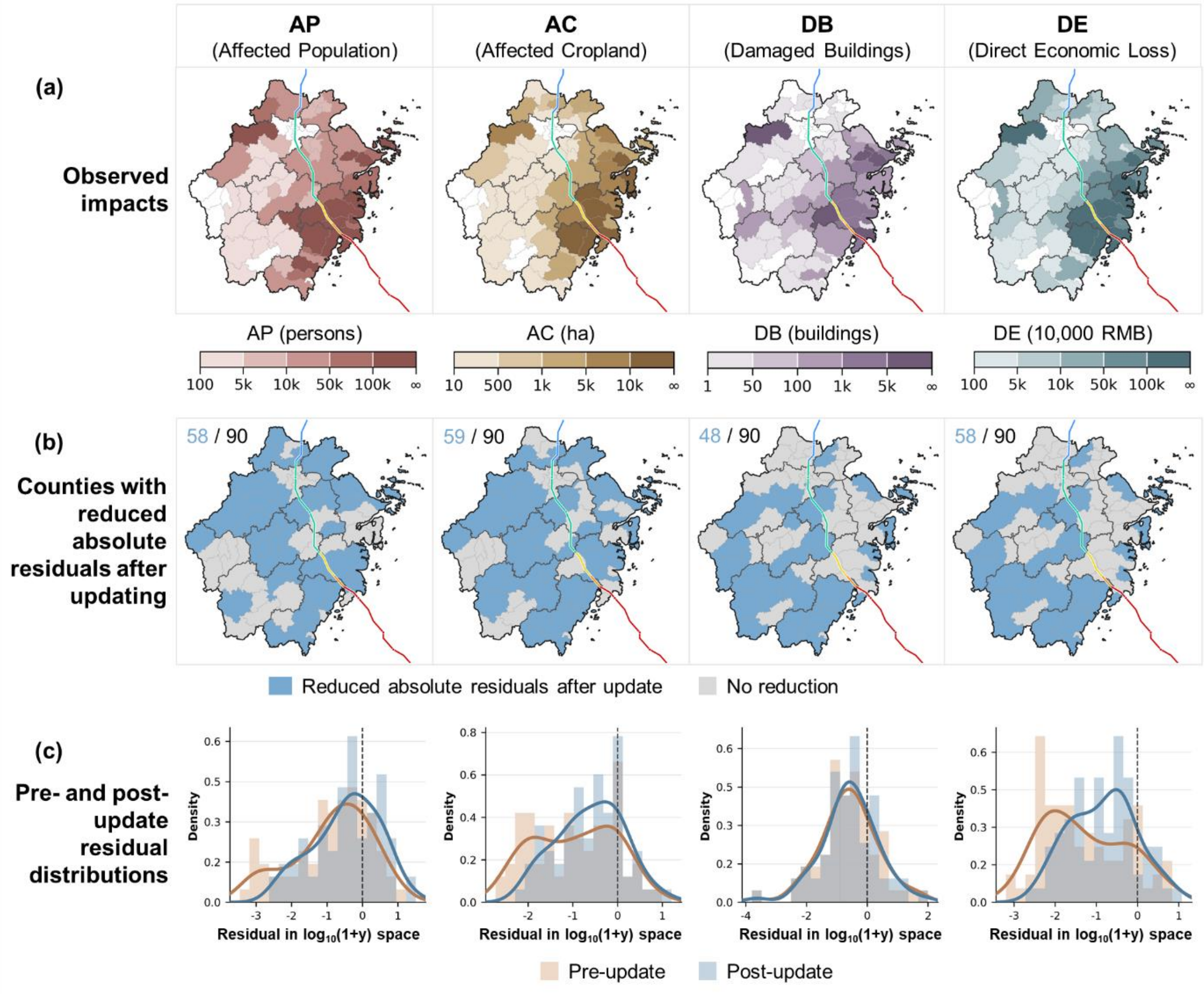

**Fig. 12.** **Cross-year updating for Typhoon Lekima (2019).** (a) Observed county-level AP, AC, DB, and DE. (b) Counties in which absolute residuals decreased after updating: 58/90 for AP, 59/90 for AC, 48/90 for DB, and 58/90 for DE. (c) Pre- and post-update residual distributions. Lekima is presented as a positive-transfer example; event-wise variability is reported in Table 7.

The ordering of these results invites an explanation, but the three protocols differ in their sources, targets, and available records. The Discussion considers source–target relevance as one interpretation while keeping the negative events and retention costs in view.

# 4. Discussion

## 4.1 Evidence utilization and system-level reliability

The 20 cyclones, rather than their nested county records, set the scale of independent event evidence. OmniCyRisk does not create new empirical storms; it organizes several related but non-interchangeable observations from each storm within one forecasting system. Fixed-evidence prediction and partial-evidence updating express this common idea in different settings: a consequence recorded for one part of the impact system may contain useful information about another, provided its distinct exposure basis and response mechanism are retained.

Marginal reliability asks whether each consequence is predicted well. Relational coherence asks whether the four forecasts together describe an impact profile resembling the observed one. Against matched Omni-STL models, OmniCyRisk produced a modest macro reliability gain, with lower DB performance, and reproduced observed pairwise consequence relationships more closely for five of six pairs. A coordinated risk vector therefore needs both checks. Good individual scores cannot by themselves show that the human, agricultural, building, and economic outputs form a credible simultaneous picture; the coherence result likewise cannot excuse weak prediction of an individual consequence.

Shared components may anchor predictions to common hazard, exposure, and vulnerability structure, while adaptive heads preserve consequence differences. The rank-correlation result is consistent with that interpretation but does not establish a causal mechanism or validate the full joint distribution. Other design choices address the limits of event evidence: simulation-assisted initialization improves the starting point without expanding the empirical archive, although its auxiliary responses inherit ISO-expert errors; exposure-consistent aggregation links fine-grid predictors to verified county totals without creating grid-level ground truth.

The appropriate model depends on the operational objective. When one consequence matters and sufficient task-specific records exist, a simpler independent model may be easier to interpret, maintain, and deploy. Coordinated forecasting becomes more attractive when several consequences must be assessed together, their consistency matters, or reporting is fragmented. The measured score and coherence differences inform this choice but do not quantify downstream decision benefit or operational cost.

## 4.2 Evidence proximity and conditional updating

Cross-consequence and cross-city updates produced substantial pooled gains, whereas cross-year gains were smaller and less consistent. This pattern motivates an evidence-proximity interpretation, although the protocols differ in their sources, targets, and available records and do not isolate proximity as a cause.

Evidence proximity means overlap in the hazard, exposure, vulnerability, and consequence-response conditions relevant to a target, rather than geographic distance or recency alone. Different outcomes of one cyclone share an event context; cities differ in local assets and vulnerability; and events in different years can differ in track, rainfall, exposure, and reporting. These differences offer a plausible interpretation of the transfer pattern. The two negative cross-year cases show why records from related hazards do not guarantee benefit.

A proposed update should be assessed before its outputs are relied on. Descriptors available at the time can help screen source–target relevance, and candidate updates can be checked on representative historical events while monitoring retained performance on prior cases. The modest reference-set decline after cross-city updating shows a stability–plasticity cost: improving a new target may erode established behavior. The seven-event post-hoc diagnostic is too small and partly depends on realized target impacts, so it does not yet provide an automatic rule for selecting sources or accepting an update. These are evaluation steps for future deployment work, not claims that this study tested an online decision system.

The backbone remained frozen, so transfer occurred through consequence-adaptive MoE heads. The controls implicate adaptable shared head branches, especially for consequence and city transfer, without isolating a unique mechanism. These experiments show recalibration of an established representation, not wholesale relearning.

In these retrospective tests, indirect records improved forecasts for withheld consequences and for non-reporting cities on average, while cross-year transfer was inconsistent. Such evidence can complement direct observations of the target consequence or place. The experiments do not determine when reports arrive, how they are revised, or when an operational user should trust an updated forecast.

### 4.3 Operational scope and broader implications

The experiments use controlled retrospective masks and realized wind and rainfall, not a complete real-time forecast chain. Reporting delays, revisions, measurement error, forecast-hazard uncertainty, and long-term shifts remain untested. The 2017 AlphaEarth layer used for pre-2017 events is an additional temporal approximation. Linear interpolation of exposure layers for missing study years may also smooth abrupt temporal changes in exposure.

All 20 cyclones and all 11 city-transfer targets lie within Zhejiang. Event blocking tests unseen cyclones, while city withholding tests updates in regions represented during base-model development. Independent regional validation remains necessary.

Global data make broader application plausible. GHSL population and built-up products can help construct comparable spatial exposure descriptions (Schiavina et al., 2023; Pesaresi and Politis, 2023), and LitPop provides distributed asset-value estimates (Eberenz et al., 2020). Vessey et al. (2026) show the value of combining hazard, exposure, and vulnerability proxies for U.S. hurricane-loss prediction; their evaluation also uses observed historical losses. For OmniCyRisk, built-up area and asset value would need to be reconciled with the building-count and GDP-based quantities used here, and none of these exposure products supplies the four independently observed consequence labels.

We therefore treat independent-region application as the next validation study. It will require harmonizing outcome definitions and event periods, assembling regional hazard and exposure layers, evaluating the framework against observed impacts in the new region, and only then testing direct transfer or local updating from Zhejiang. A forecast map made with global exposures alone would illustrate portability but would not establish external predictive skill (Supplementary Table S8). We will pursue this broader test in subsequent work; the present empirical claims remain within Zhejiang.

The 1-km fields remain latent allocations, and pairwise coherence is a limited description of dependence. ISO comparisons test the whole framework, whereas Omni-STL provides the architecture-matched sharing comparison. Finer-resolution observations, external regions, and richer joint-distribution tests would address these limits.

Within the tested region, OmniCyRisk offers a framework for four coordinated impact forecasts and for updating unobserved targets from partial reports, with diagnostics that show where those capabilities help or fail. A prospective rule for selecting or accepting updates and a decision-level cost–benefit analysis remain to be developed.

## 5. Conclusion

OmniCyRisk treats scarce tropical cyclone impact records as connected but non-interchangeable evidence for four outcomes. Its shared spatial representation and consequence-adaptive heads coordinate forecasts of affected population, affected cropland, damaged buildings, and direct economic loss. Exposure-consistent aggregation anchors fine-grid outputs to observed county totals; simulation-assisted initialization supplies a stronger starting point under the tested refinement protocol without augmenting the observed event archive; and masked updating tests what reports from other consequences, cities, or years can contribute to targets excluded from the update.

In event-blocked tests, OmniCyRisk was competitive with architecture-matched Omni-STL models (macro reliability 0.685 versus 0.675), although damaged-building reliability was lower. Its forecasts reproduced observed pairwise rank relationships more closely for five of six consequence pairs, reducing mean correlation error by 59.4% descriptively; simulation-assisted initialization improved 19 of 20 consequence–fold comparisons. Controlled partial-evidence updates improved every withheld consequence and produced positive skill point estimates for all 11 non-reporting cities within Zhejiang. These gains were not cost-free: several city intervals included zero, and some prior-reference scores declined. Cross-year updating showed smaller, less consistent gains: pooled skill was positive and five of seven cyclones had positive point estimates, while Hinnamnor and Muifa had negative point estimates.

The wider insight is that scarcity of independent events is also an evidence-organization problem. Reports can carry useful information across consequences and places when the system shares relevant structure while retaining each outcome's exposure basis and response mechanism. These findings point toward multi-sector impact forecasts that can incorporate partial reports for still-unreported targets, with checks for negative transfer and retained performance. The present tests are retrospective, use realized hazard fields, and remain within Zhejiang. Independent regional archives, forecast-hazard evaluation, finer-resolution impact observations, and prospective update and decision studies are needed to determine how broadly the approach transfers and when it improves real-world decisions. Within its tested setting, OmniCyRisk shows that fragmented impact records can yield a more coherent multi-consequence forecast and improve predictions for targets absent from an update.

## Acknowledgements

This project is supported by National Natural Science Foundation of China through Grant No. 52578613.

# Supplementary Material

## Supplementary Methods

### S1. Simulation-assisted initialization procedure

Simulation-assisted initialization was used to expose OmniCyRisk to a broader range of plausible tropical cyclone hazard–impact conditions before refinement with the limited historical impact records. The overall procedure consisted of synthetic tropical cyclone generation, parametric wind and rainfall simulation, matching of environmental and exposure conditions, generation of auxiliary impact responses, and county-level initialization of OmniCyRisk.

#### S1.1 Synthetic tropical cyclone and hazard-field generation

The synthetic tropical cyclone tracks were provided by WindRiskTech LLC and generated using a statistical–deterministic tropical cyclone simulation framework following Emanuel et al. (2006, 2008). The stochastic catalog represented approximately 80 years of tropical cyclone activity affecting Zhejiang Province. A total of 240 synthetic events influencing the study region were retained to construct the simulation library.

For each synthetic event, the time-varying cyclone track and storm parameters were converted into gridded wind and rainfall fields. Tropical cyclone track and intensity evolution were represented following Chen and Duan (2018). Near-surface wind fields were generated using the parametric wind-field formulation of Georgiou et al. (1983), while rainfall fields were simulated using the parametric rainfall model described by Geoghegan et al. (2018).

The hourly wind and rainfall fields were summarized using the same hazard definitions adopted for the historical events: maximum hourly 10-min mean wind speed (MW), maximum hourly rainfall (MR), and maximum 12-h accumulated rainfall (AR). All hazard fields were aligned to the common 1-km modeling grid.

#### S1.2 Matching of environmental and exposure conditions

The synthetic wind and rainfall fields describe event-specific hazard forcing but do not specify the environmental and exposure conditions associated with each event. Therefore, one year between 2012 and 2022 was randomly assigned to each synthetic tropical cyclone.

The assigned year determined the background conditions paired with each synthetic event. Exposure layers followed the same source-specific temporal matching used for the historical events, whereas AlphaEarth Embeddings followed the study-wide temporal rule: the 2017 AEE layer was used for assigned years from 2012 to 2016, and the corresponding annual AEE layer was used from 2017 onward. The resulting background predictors comprised 64 AlphaEarth embedding channels together with population, cropland, building, and GDP exposure layers.

Each synthetic event consequently had the same 71-channel input structure as the historical events, comprising 3 hazard variables, 64 environmental channels, and 4 exposure variables.

### S1.3 Generation of auxiliary impact responses

Because the synthetic scenarios had no observed disaster-impact records, auxiliary responses were generated using the four isolated consequence-specific experts: ISO-AP, ISO-AC, ISO-DB, and ISO-DE.

For each experimental training set, the four experts were trained exclusively using the corresponding historical impact records. Validation records were used only for model selection and early stopping, while test records were excluded from model development. The selected experts were then applied to the 240 synthetic scenarios to estimate affected population (AP), affected cropland (AC), damaged buildings (DB), and direct economic loss (DE).

For each consequence, the expert first generated a grid-level impact-ratio field. The predicted ratios were multiplied by the corresponding exposure layer and aggregated within county boundaries to obtain county-level auxiliary impact responses. The resulting simulated predictors and four county-level auxiliary responses formed the simulation library used for OmniCyRisk initialization. These expert-generated responses were treated as model-based auxiliary supervision rather than as additional empirical observations.

### S1.4 County-level initialization of OmniCyRisk

During simulation-assisted initialization, the complete OmniCyRisk model was trained using the synthetic predictor fields and the county-level auxiliary responses generated by the isolated experts. OmniCyRisk produced grid-level impact-ratio fields, which were converted into county-level predictions using the same exposure-consistent aggregation procedure adopted for the historical observations.

The initialization loss was calculated only at the county level. Grid-level outputs from the isolated experts were not directly used as supervision targets because the experts were trained and evaluated using county-level impact records, whereas their internal grid-level ratio fields remained latent and were not independently validated. County-level supervision therefore maintained consistency with the aggregate weak-supervision setting of the overall framework.

Simulation-assisted initialization established a structured initial parameter state for the shared backbone and consequence-adaptive prediction heads. The initialized model was subsequently refined using observed multi-consequence impact records. Synthetic responses were not used for final calibration, model selection, or held-out evaluation; their role was restricted to improving the initial optimization state of OmniCyRisk.

## S2. Detailed optimization and update-skill definitions

The main-text Methods describe the training sequence; the equations below provide the exact loss, reliability, and update-skill definitions used in the analyses.

### S2.1 Asymmetric consequence loss and uncertainty weighting

Impact records are strongly skewed and span several orders of magnitude. Observed and predicted county impacts are first transformed as $\log_{10}(1 + \text{impact})$; Eq. (S4) gives the formal notation. For each consequence, the objective is an asymmetric mean squared error over its available training records:

$$L_k = \frac{1}{N_k}\sum_{i=1}^{N_k} \omega_i^{(k)} \left(\hat{z}_i^{(k)} - z_i^{(k)}\right)^2 \tag{S1}$$

The observation-specific weight is:

$$\omega_i^{(k)} = \begin{cases} \alpha, & \hat{z}_i^{(k)} < z_i^{(k)}, \\ \beta, & \hat{z}_i^{(k)} \geq z_i^{(k)}. \end{cases} \tag{S2}$$

We set $\alpha = 5$ and $\beta = 1$, assigning five times as much weight to underestimation as to overestimation of the same transformed magnitude. This asymmetry reflects the operational cost of insufficient impact estimates while retaining a continuous, magnitude-sensitive objective. Each ISO expert minimizes the corresponding $L_k$.

OmniCyRisk jointly optimizes the four consequence-level losses using a softplus-regularized uncertainty-weighting objective adapted from Kendall et al. (2018):

$$L_{\text{Omni}} = \sum_{k=1}^{K} \left[ \frac{1}{2} \exp(-s_k) L_k + \frac{1}{2} \log\big(1 + \exp(s_k)\big) \right] \tag{S3}$$

where $s_k$ is a learnable consequence-specific uncertainty parameter. The first term scales each loss according to its learned uncertainty, while the softplus term prevents unbounded attenuation through arbitrarily large $s_k$. The $s_k$ parameters are optimized with the complete model during simulation-assisted initialization and with the consequence-adaptive heads during empirical refinement.

## S2.2 Reliability components and normalized skill

For observed and predicted county-level impacts, the log transformation used in optimization and evaluation is:

$$z_i^{(k)} = \log_{10}\left(1 + y_i^{(k)}\right), \qquad \hat{z}_{i,F}^{(k)} = \log_{10}\left(1 + \hat{y}_{i,F}^{(k)}\right) \tag{S4}$$

After the validation-selected impact threshold classifies predictions, hits, misses, and false alarms define the critical success index:

$$CSI_k = \frac{H_k}{H_k + M_k + FA_k} \tag{S5}$$

Positive-impact order-of-magnitude agreement uses the stated one-order-of-magnitude tolerance:

$$OMA_{k,+}(\delta) = \frac{1}{N_{k,+}} \sum_{i \in \mathcal{I}_{k,+}} I\left(\left|\hat{z}_{i,F}^{(k)} - z_i^{(k)}\right| \leq \delta\right), \quad \delta = 1 \tag{S6}$$

Underestimation avoidance penalizes the accumulated shortfall below observed impacts:

$$UAS_{k,+} = 1 - \frac{\sum_{i \in \mathcal{I}_{k,+}} \max\left(0, z_i^{(k)} - \hat{z}_{i,F}^{(k)}\right)}{\sum_{i \in \mathcal{I}_{k,+}} z_i^{(k)}} \tag{S7}$$

The macro score averages the four consequence-specific geometric-mean scores:

$$\overline{P}^{\text{risk}}(F; \mathcal{T}) = \frac{1}{K} \sum_{k=1}^{K} P_k^{\text{risk}}(F; \mathcal{T}) \tag{S8}$$

For a specified comparison model, normalized skill is calculated for each consequence as:

$$S_k^{\text{risk}}(\mathcal{T}) = \frac{P_k^{\text{risk}}(F_{\text{Omni}}; \mathcal{T}) - P_k^{\text{risk}}(F_{\text{iso}}; \mathcal{T})}{1 - P_k^{\text{risk}}(F_{\text{iso}}; \mathcal{T})} \tag{S9}$$

The corresponding macro normalized skill uses the two macro scores:

$$\overline{S}^{\text{risk}}(\mathcal{T}) = \frac{\overline{P}^{\text{risk}}(F_{\text{Omni}};\mathcal{T}) - \overline{P}^{\text{risk}}(F_{\text{iso}};\mathcal{T})}{1 - \overline{P}^{\text{risk}}(F_{\text{iso}};\mathcal{T})} \tag{S10}$$

These equations reproduce the definitions used for all reported comparisons. The macro score and the pooled score differ when target sizes and baseline headroom differ; target-equal and pooled update skill are defined separately in Eqs. (S11) and (S12).

### S2.3 Target-equal and pooled update skill

The reliability score combines affected-location detection, positive-impact magnitude agreement, and underestimation avoidance; its exact components follow in Section S2.2. For an excluded target with pre-update and post-update scores, target-equal macro skill averages normalized skill across targets. Pooled skill first combines held-out evaluation records, or the four consequence scores in the cross-consequence experiment, and then normalizes the pre-to-post improvement by the remaining headroom.

$$\overline{S}^{\text{upd}}_{\text{average}} = \frac{1}{|\mathcal{H}|}\sum_{h\in\mathcal{H}} S_h^{\text{upd}} \tag{S11}$$

$$S^{\text{upd}}_{\text{pool}} = \frac{P^{\text{risk}}_{\mathcal{H},\text{post}} - P^{\text{risk}}_{\mathcal{H},\text{pre}}}{1 - P^{\text{risk}}_{\mathcal{H},\text{pre}}}. \tag{S12}$$

The target-equal and pooled quantities differ when target sizes and baseline headroom vary. They are labeled separately in Tables 5–7.

## S3. County-group definitions for the spatial sensitivity analysis

Coastal counties were defined by centroid-to-coast distance below 40 km (43 coastal and 47 non-coastal counties; 860 and 940 county–event records). Mountainous counties had mean elevation above 200 m (23 mountainous and 67 non-mountainous; 460 and 1,340 records). Population groups used the county median density (45 dense and 45 non-dense; 900 records each). These within-province strata describe heterogeneity and are not independent tests of geographic generalization.

## Supplementary Tables

**Table S1. Historical Tropical Cyclone Events and Event-Blocked Cross-Validation Assignments**

| No. | TC ID | TC Name | Start time | End time | Duration (h) | Max. wind (m/s) | Assigned fold | Severity rank |
|---|---|---|---|---|---|---|---|---|
| 1 | 201211 | Haikui | 2012-08-05 12:00 | 2012-08-09 12:00 | 97 | 45 | 3 | 3 |
| 2 | 201307 | Soulik | 2013-07-12 04:00 | 2013-07-14 19:00 | 64 | 42 | 2 | 14 |
| 3 | 201312 | Trami | 2013-08-20 15:00 | 2013-08-22 22:00 | 56 | 35 | 5 | 11 |
| 4 | 201323 | Fitow | 2013-10-05 13:00 | 2013-10-07 13:00 | 49 | 45 | 1 | 1 |
| 5 | 201410 | Matmo | 2014-07-22 03:00 | 2014-07-25 14:00 | 84 | 42 | 4 | 17 |
| 6 | 201509 | Chan-hom | 2015-07-10 04:00 | 2015-07-12 19:00 | 64 | 55 | 5 | 5 |
| 7 | 201513 | Soudelor | 2015-08-07 10:00 | 2015-08-10 23:00 | 86 | 50 | 4 | 4 |
| 8 | 201614 | Meranti | 2016-09-13 15:00 | 2016-09-17 04:00 | 86 | 75 | 2 | 9 |
| 9 | 201617 | Megi | 2016-09-27 00:00 | 2016-09-29 23:00 | 72 | 52 | 3 | 8 |
| 10 | 201808 | Maria | 2018-07-09 21:00 | 2018-07-12 07:00 | 59 | 52 | 3 | 13 |
| 11 | 201810 | Ampil | 2018-07-19 09:00 | 2018-07-23 09:00 | 97 | 28 | 1 | 20 |
| 12 | 201812 | Jongdari | 2018-08-01 00:00 | 2018-08-04 12:00 | 85 | 23 | 5 | 16 |
| 13 | 201814 | Yagi | 2018-08-10 07:00 | 2018-08-14 07:00 | 97 | 28 | 2 | 19 |
| 14 | 201909 | Lekima | 2019-08-07 17:00 | 2019-08-11 17:00 | 97 | 62 | 2 | 2 |
| 15 | 201918 | Mitag | 2019-09-28 21:00 | 2019-10-02 21:00 | 97 | 40 | 1 | 10 |

| No. | TC ID | TC Name | Start time | End time | Duration (h) | Max. wind (m/s) | Assigned fold | Severity rank |
|---|---|---|---|---|---|---|---|---|
| 16 | 202004 | Hagupit | 2020-08-01 07:00 | 2020-08-05 07:00 | 97 | 42 | 5 | 6 |
| 17 | 202106 | In-Fa | 2021-07-23 09:00 | 2021-07-27 09:00 | 97 | 38 | 4 | 7 |
| 18 | 202114 | Chanthu | 2021-09-10 14:00 | 2021-09-14 14:00 | 97 | 68 | 1 | 15 |
| 19 | 202211 | Hinnamnor | 2022-09-02 13:00 | 2022-09-06 08:00 | 92 | 55 | 3 | 18 |
| 20 | 202212 | Muifa | 2022-09-11 21:00 | 2022-09-15 21:00 | 97 | 50 | 4 | 12 |

**Note:** All start and end times are reported in UTC+08:00. Duration denotes the number of hourly time steps within the selected analysis window, including both endpoints. Maximum wind speed denotes the highest 2-min mean sustained wind speed near the cyclone center recorded in the CMA tropical cyclone best-track dataset during the corresponding analysis window. Assigned fold indicates the event allocation for the five-fold event-blocked cross-validation described in Section 2.4.1, with all county-level records from the same tropical cyclone retained within a single fold. Severity rank was determined by averaging the event-level severity across affected population (AP), affected cropland (AC), damaged buildings (DB), and direct economic loss (DE), and then ranking the 20 events in descending order; rank 1 represents the most severe event.

**Table S2. Detailed Architecture and Training Configuration of OmniCyRisk and the Isolated Consequence-Specific Experts**

**Table S2a.** Architecture and Stage-Specific Training Configuration of OmniCyRisk

**Table S2a-1.** Architectural Configuration of OmniCyRisk

| Category | Item | Configuration |
|---|---|---|
| **Input** | Predictors | 71-channel hazard, environmental, and exposure predictors |
| | Spatial dimensions | (540×540) grid cells at 1-km resolution |
| **Patch embedding** | Input projection | (5×5) convolution with stride 5, GroupNorm, and GELU |
| | Auxiliary skip branch | (3×3) convolution with stride 2 |
| | Positional encoding | Learnable positional embeddings added to patch embeddings |
| **Encoder** | Encoder architecture | Three-stage hierarchical Swin Transformer |
| | Feature dimensions | 64, 128, and 256 |
| | Block depths | 2, 2, and 2 |
| | Attention heads | 2, 4, and 8 |
| | Window size | 6 |

| Category | Item | Configuration |
|---|---|---|
| **Decoder** | Downsampling | Patch merging |
| | Configuration | Three-stage CNN-based decoder comprising bilinear upsampling, skip-feature concatenation, and 3×3 convolution |
| | Feature dimensions | 128, 64, and 32 |
| **Refinement** | Configuration | Three (3×3) convolutional blocks with 32 features |
| **Prediction heads** | Head type | Four consequence-adaptive mixture-of-experts heads |
| | Expert branches | Two shared and one consequence-specific expert per head |
| | Expert configuration | (3×3) convolution, 32→16 features, GroupNorm, and GELU |
| | Gating mechanism | Input-dependent softmax gating over the three expert branches |
| **Output** | Output layer | Sigmoid activation producing four grid-level impact-ratio fields |
| **Regularization** | Normalization / activation | GroupNorm and LayerNorm / GELU |
| | Dropout rate | 0.05 |
| **Capacity** | Total parameters | 3,210,284 |

**Note:** The four consequence-adaptive MoE heads correspond to affected population (AP), affected cropland (AC), damaged buildings (DB), and direct economic loss (DE). Each head combines shared and consequence-specific expert branches through learned softmax gating. The predicted grid-level impact ratios are converted into county-level absolute impacts using the corresponding exposure layer.

**Table S2a-2.** Stage-Specific Training Configuration of OmniCyRisk

| Item | Simulation-assisted initialization | Empirical refinement | Controlled retrospective updating |
|---|---|---|---|
| **Training data** | 240 simulated TC scenarios with ISO-expert-generated auxiliary responses | Observed multi-consequence records from the fold-specific training events | Partially observed post-2018 records under event, region, or consequence masks |
| **Trainable modules** | All model parameters | Only consequence-adaptive MoE heads | Only consequence-adaptive MoE heads |
| **Loss function** | Asymmetric MSE in transformed impact space | Same | Same |
| **Under-/overestimation weights** | 5.0 / 1.0 | Same | Same |
| **Consequence weighting** | Learnable log-variance parameters constrained to ([-3,3]) | Same | Same |
| **Optimizer** | AdamW | Same | Same |
| **Initial learning rate** | $5\times10^{-4}$ | $5\times10^{-5}$ | $5\times10^{-5}$ |
| **Minimum learning rate** | $1\times10^{-6}$ | $1\times10^{-7}$ | $1\times10^{-7}$ |

| Item | Simulation-assisted initialization | Empirical refinement | Controlled retrospective updating |
|---|---|---|---|
| **Weight decay** | $1\times10^{-6}$ | $1\times10^{-4}$ | $1\times10^{-4}$ |
| **Batch size** | 8 | 2 | 2 |
| **Learning-rate scheduler** | ReduceLROnPlateau | Same | Same |
| **Scheduler factor / patience** | 0.8 / 15 epochs | 0.5 / 5 epochs | 0.5 / 5 epochs |
| **Early-stopping patience** | 50 epochs | 25 epochs | 25 epochs |
| **Model selection** | Minimum validation loss | Same | Same |

Note: Simulation-assisted initialization optimizes the complete OmniCyRisk model using expert-generated auxiliary responses. During empirical refinement and controlled retrospective updating, the shared backbone is frozen and only the consequence-adaptive MoE heads are updated. The reported random-initialization ablation follows the same refinement protocol from random starting parameters; it is not an end-to-end observed-data fit with a trainable backbone throughout.

**Table S2b.** Architecture and Training Configuration of the Isolated Consequence-Specific Experts

| Category | Item | Configuration |
|---|---|---|
| **Model** | Expert models | ISO-AP, ISO-AC, ISO-DB, and ISO-DE |
| | Architecture | Independently trained consequence-specific CNNs with an identical encoder–decoder architecture |
| **Input** | Predictors | 71-channel hazard, environmental, and exposure predictors |
| **Encoder** | Convolutional stages | Three stages with feature dimensions of 32, 64, and 128 |
| | Basic block | (3×3) convolution, BatchNorm, GELU, and max pooling |
| **Bottleneck** | Configuration | (3×3) convolution with 128 features and Dropout2d |
| **Decoder** | Configuration | Bilinear upsampling with encoder–decoder skip connections |
| | Feature dimensions | 64, 32, and 16 |
| **Output** | Output layer | (1×1) convolution followed by Sigmoid |
| | Prediction | One grid-level impact-ratio field per expert |
| **Training** | Trainable parameters | All model parameters |
| | Loss function | Asymmetric MSE in transformed impact space |
| | Under-/overestimation weights | 5.0 / 1.0 |
| | Optimizer | AdamW |
| | Initial learning rate | $5\times10^{-4}$ |
| | Weight decay | $1\times10^{-5}$ |
| | Batch size | 2 |
| | Learning-rate scheduler | ReduceLROnPlateau |
| | Scheduler factor / patience | 0.8 / 10 epochs |
| | Early-stopping patience | 25 epochs |
| **Capacity** | Parameters per expert | 455,009 |

**Note:** The four isolated consequence-specific experts use the same predictors, network architecture, output definition, and exposure-consistent aggregation procedure, but are trained independently using records of the corresponding consequence. AP, AC, DB, and DE denote affected population, affected cropland, damaged buildings, and direct

economic loss, respectively.

**Table S3. Fold-wise performance comparison between OmniCyRisk and isolated consequence-specific experts under event-blocked cross-validation**

| Consequence | Fold 1 | Fold 2 | Fold 3 | Fold 4 | Fold 5 | Win count |
|---|---|---|---|---|---|---|
| AP | + | - | + | + | + | **4/5** |
| AC | + | + | + | + | - | **4/5** |
| DB | + | + | + | - | + | **4/5** |
| DE | + | + | + | + | + | **5/5** |
| Overall | + | + | + | + | + | **17/20** |

**Note:** AP, AC, DB, and DE denote affected population, affected cropland, damaged buildings, and direct economic loss, respectively. A "+" indicates that OmniCyRisk achieved a higher $P_k^{\text{risk}}$ than the corresponding isolated consequence-specific expert in the same held-out fold, whereas a "–" indicates lower performance. The win count summarizes the number of folds in which OmniCyRisk outperformed the corresponding isolated expert across the five event-blocked folds. This comparison provides a framework-level reference against isolated consequence-specific models and does not isolate the contribution of cross-consequence sharing.

**Table S4. Fold-wise and pooled reliability comparison between architecture-matched OmniCyRisk and Omni-STL under event-blocked cross-validation**

**Table S4a.** Fold-wise composite reliability ($P_k^{\text{risk}}$) comparison between OmniCyRisk and Omni-STL

| Consequence | Model | Fold 1 | Fold 2 | Fold 3 | Fold 4 | Fold 5 | Win count |
|---|---|---|---|---|---|---|---|
| AP | OmniCyRisk | **0.775** | **0.724** | **0.806** | **0.695** | **0.708** | **5/5** |
| | Omni-STL | 0.7226 | 0.720 | 0.752 | 0.686 | 0.660 | |
| AC | OmniCyRisk | **0.834** | **0.593** | **0.764** | **0.540** | 0.714 | **4/5** |
| | Omni-STL | 0.833 | 0.575 | 0.652 | 0.521 | **0.725** | |
| DB | OmniCyRisk | 0.659 | **0.625** | **0.724** | 0.523 | 0.710 | **2/5** |
| | Omni-STL | **0.670** | 0.596 | 0.723 | **0.599** | **0.733** | |
| DE | OmniCyRisk | **0.760** | **0.569** | **0.756** | 0.650 | 0.661 | **3/5** |
| | Omni-STL | 0.759 | 0.550 | 0.737 | **0.655** | **0.682** | |

**Table S4b.** Pooled out-of-fold reliability components of OmniCyRisk and Omni-STL

| Consequence | Model | CSI | OMA₊ | UAS₊ |
|---|---|---|---|---|
| AP | OmniCyRisk | **0.582 [0.551, 0.615]** | **0.782 [0.754, 0.811]** | **0.891 [0.880, 0.903]** |
| | Omni-STL | 0.537 [0.507, 0.567] | 0.748 [0.717, 0.780] | 0.878 [0.865, 0.891] |
| AC | OmniCyRisk | **0.538 [0.501, 0.575]** | **0.711 [0.673, 0.752]** | **0.849 [0.832, 0.867]** |
| | Omni-STL | 0.484 [0.451, 0.520] | 0.706 [0.667, 0.749] | 0.837 [0.817, 0.855] |
| DB | OmniCyRisk | 0.391 [0.357, 0.427] | 0.762 [0.722, 0.803] | **0.844 [0.821, 0.868]** |
| | Omni-STL | **0.427 [0.392, 0.463]** | **0.806 [0.770, 0.843]** | 0.828 [0.804, 0.850] |
| DE | OmniCyRisk | 0.564 [0.525, 0.604] | **0.644 [0.607, 0.681]** | **0.875 [0.860, 0.888]** |
| | Omni-STL | 0.567 [0.527, 0.610] | 0.631 [0.594, 0.666] | 0.873 [0.858, 0.887] |
| **Macro average** | OmniCyRisk | **0.519 [0.492, 0.546]** | **0.725 [0.703, 0.747]** | **0.865 [0.853, 0.877]** |
| | Omni-STL | 0.504 [0.478, 0.531] | 0.723 [0.701, 0.745] | 0.854 [0.842, 0.866] |

**Note:** AP, AC, DB, and DE denote affected population, affected cropland, damaged buildings, and direct economic loss, respectively. Omni-STL denotes the architecture-matched single-task counterpart of OmniCyRisk, consisting of four independently trained consequence-specific models. Each model uses the same backbone architecture, prediction pathway, initialization strategy, optimization settings, aggregation procedure, and evaluation protocol as OmniCyRisk, but optimizes one consequence independently without cross-consequence sharing. Table S4a reports fold-wise composite reliability ($P_k^{\mathrm{risk}}$) across five event-blocked folds. Table S4b reports component-level reliability metrics calculated from pooled out-of-fold predictions across five event-blocked folds. Brackets indicate 95% bootstrap confidence intervals. Bold values indicate the higher performance between OmniCyRisk and Omni-STL for the same consequence and metric.

## **Table S5.** Control Experiments and Reference-Set Performance Retention during Updating with Partially Observed Evidence

**Table S5a.** Consequence-wise additional skill of full updating over the frozen-shared-expert control and reference-set performance change

| Held-out consequence | Additional skill of full updating over the frozen-shared-expert control (%) [95% CI] | Pre-2019 reference-set change, $\Delta P_{\mathrm{k}}^{\mathrm{risk}}$ |
|---|---|---|
| **AP** | **23.9** [14.6, 33.7] | -0.006 [-0.046, 0.036] |
| **AC** | **37.1** [24.9, 41.6] | -0.048 [-0.074, -0.016] |

| Held-out consequence | Additional skill of full updating over the frozen-shared-expert control (%) [95% CI] | Pre-2019 reference-set change, $\Delta P_{k}^{risk}$ |
|---|---|---|
| **DB** | **16.7** [5.2, 22.5] | **0.029** [-0.010, 0.103] |
| **DE** | **31.1** [22.6, 39.0] | -0.032 [-0.065, -0.006] |
| **Macro average** | **27.2** [21.7, 29.9] | -0.014 [-0.029, 0.010] |
| **Pooled** | **26.3** [20.1, 28.8] | -0.014 [-0.029, 0.010] |

**Table S5b.** Event-wise additional skill of full updating over the frozen-shared-expert control and reference-set performance change

| Held-out event | Additional skill of full updating over the frozen-shared-expert control (%) [95% CI] | Pre-2019 reference-set change, $\Delta \overline{P}^{risk}$ |
|---|---|---|
| **Lekima (2019)** | **6.3** | -0.001 [-0.039, 0.035] |
| **Mitag (2019)** | **0.8** | - |
| **Hagupit (2020)** | **9.9** | -0.006 [-0.042, 0.031] |
| **In-Fa (2021)** | **3.1** | **0.002** [-0.032, 0.040] |
| **Chanthu (2021)** | -0.8 | - |
| **Hinnamnor (2022)** | -2.5 | **0.014** [-0.008, 0.036] |
| **Muifa (2022)** | -3.9 | - |
| **Macro average** | **1.8** | **0.003** [-0.001, 0.009] |
| **Pooled** | **3.6** [-1.4, 5.6] | **0.003** [-0.002, 0.009] |

**Table S5c.** City-wise additional skill of full updating over the frozen-shared-expert control and reference-set performance change

| Held-out city | Additional skill of full updating over the frozen-shared-expert control (%) [95% CI] | Pre-2019 reference-set change, $\Delta \overline{P}^{risk}$ |
|---|---|---|
| **Coastal cities** | | |
| Wenzhou (WZ) | 34.1 [11.5, 49.1] | -0.036 [-0.063, -0.009] |
| Taizhou (TZ) | 0.5 [-4.5, 5.9] | -0.007 [-0.020, 0.005] |
| Ningbo (NB) | 0.8 [-13.0, 13.8] | -0.060 [-0.088, -0.033] |
| Zhoushan (ZS) | -3.1 [-12.2, 5.0] | -0.003 [-0.013, 0.007] |
| Jiaxing (JX) | 31.9 [-6.7, 46.2] | -0.042 [-0.068, -0.017] |
| **Coastal macro average** | **12.9** [2.8, 17.8] | **-0.030** [-0.040, -0.019] |

| Held-out city | Additional skill of full updating over the frozen-shared-expert control (%) [95% CI] | Pre-2019 reference-set change, $\Delta\overline{P}^{\text{risk}}$ |
|---|---|---|
| **Inland cities** | | |
| Hangzhou (HZ) | 24.6 [14.2, 35.3] | -0.039 [-0.063, -0.012] |
| Huzhou (HU) | 18.1 [-10.8, 40.9] | -0.037 [-0.063, -0.013] |
| Shaoxing (SX) | 31.2 [17.2, 43.6] | -0.052 [-0.080, -0.023] |
| Jinhua (JH) | 48.3 [18.2, 73.1] | -0.046 [-0.072, -0.018] |
| Quzhou (QZ) | 32.9 [-16.2, 58.1] | -0.029 [-0.055, -0.003] |
| Lishui (LS) | 57.8 [14.9, 78.3] | -0.047 [-0.075, -0.022] |
| **Inland macro average** | **35.5** [22.6, 45.6] | **-0.042** [-0.053, -0.031] |
| **Macro average** | **25.2** [16.1, 30.7] | **-0.036** [-0.044, -0.029] |
| **Pooled** | **20.8** [14.6, 26.6] | **-0.037** [-0.045, -0.030] |

Note: Tables S5a and S5c report 95% record-level bootstrap intervals. Table S5b event rows and the macro average are point estimates; only the pooled estimate has an interval. Reference-set change is the post-update minus pre-update composite score, so negative values indicate loss of prior performance. All targets were excluded from updating. The shared backbone was frozen; the control additionally froze the shared expert branches within the MoE heads. For years with multiple held-out cyclones, reference-set change is reported once because the same year-level updated model applies to all events in that year.

**Table S6. Post-hoc source–target relevance diagnostics associated with cross-year transfer heterogeneity**

| Held-out event | Normalized update skill (%) | Hazard descriptor distance | Impact-profile distance |
|---|---|---|---|
| Lekima (2019) | **10.4** | 3.45 | 2.32 |
| Mitag (2019) | **5.9** | 1.18 | 0.68 |
| Hagupit (2020) | **18.9** | 2.09 | 0.62 |
| In-Fa (2021) | **6.0** | 1.22 | 0.57 |
| Chanthu (2021) | **3.8** | 2.16 | 1.14 |
| Hinnamnor (2022) | -2.6 | 2.94 | 3.04 |
| Muifa (2022) | -1.7 | 1.18 | 0.73 |

**Note:** Hazard descriptor distance was calculated as the Euclidean distance between z-standardized event-level hazard descriptors. Impact-profile distance was calculated as the Euclidean distance between z-standardized four-consequence impact profiles, including affected population, affected cropland, damaged buildings, and direct economic loss. Smaller distances indicate greater source–target similarity. These post-hoc diagnostics were used only for interpretation and were not involved in model updating, source selection, or evaluation.

**Table S7. City-wise cross-city updating performance for all 11 held-out prefecture-level cities in Zhejiang Province**

| Held-out city | Pre-update $\overline{P}^{\text{risk}}$ | Post-update $\overline{P}^{\text{risk}}$ | Normalized update skill (%) [95% CI] |
|---|---|---|---|
| **Coastal cities** | | | |
| Wenzhou (WZ) | 0.652 | 0.782 | **37.5** [15.0, 52.3] |
| Taizhou (TZ) | 0.659 | 0.667 | **2.2** [-3.4, 8.5] |
| Ningbo (NB) | 0.792 | 0.796 | **2.2** [-14.3, 15.1] |
| Zhoushan (ZS) | 0.661 | 0.693 | **9.4** [1.2, 18.0] |
| Jiaxing (JX) | 0.506 | 0.719 | **41.3** [10.1, 58.2] |
| **Coastal macro average** | **0.654** | **0.730** | **18.5** [10.9, 23.8] |
| **Inland cities** | | | |
| Hangzhou (HZ) | 0.345 | 0.521 | **26.9** [17.2, 36.8] |
| Huzhou (HU) | 0.468 | 0.599 | **24.6** [-6.4, 48.7] |
| Shaoxing (SX) | 0.501 | 0.662 | **32.2** [18.8, 43.1] |
| Jinhua (JH) | 0.475 | 0.746 | **51.7** [21.5, 75.5] |
| Quzhou (QZ) | 0.462 | 0.640 | **32.9** [-16.2, 58.1] |
| Lishui (LS) | 0.606 | 0.839 | **59.2** [24.3, 78.5] |
| **Inland macro average** | **0.476** | **0.668** | **37.9** [25.5, 48.0] |
| **Macro average** | **0.557** | **0.696** | **29.1** [20.7, 34.5] |
| **Pooled** | **0.617** | **0.712** | **24.9** [18.7, 30.5] |

**Note:** Each city was excluded from updating and model selection in its own retrospective experiment. Positive normalized update skill favors updating. City-wise confidence intervals include zero for some locations, so positive point estimates should not be read as individually conclusive gains. Coastal and inland macro averages weight their cities equally; the pooled estimate aggregates held-out records before normalization.

**Table S8. Candidate data components and validation requirements for an independent-region extension**

| Consequence | Candidate external exposure | Requirement for evaluation |
|---|---|---|
| Affected population | GHSL population count or locally harmonized population grid | Observed affected-person totals with matching event and administrative definitions. |
| Affected cropland | A temporally matched cropland-area layer | Observed affected-cropland area and a compatible definition of agricultural impact. |
| Damaged buildings | GHSL built-up information or building footprints, converted to a compatible count basis | Observed damaged-building counts and a validated count conversion; built-up area is not itself a damaged-building count. |
| Direct economic loss | LitPop asset value or a regionally comparable economic-exposure layer | Observed direct-loss totals and a reconciled exposure definition; asset value is not identical to the GDP basis used here. |

**Note:** These are candidate inputs for future work, not datasets substituted into the current model or an external-validation result. Every regional evaluation also requires event-specific hazard inputs, consistent administrative boundaries, and independent observed impact records. Dataset sources and the validation sequence are discussed in Section 4.3 of the main manuscript.